%% file: example_paper.tex
\documentclass{article}

\usepackage{microtype}
\usepackage{graphicx}
\usepackage{subcaption}
\usepackage{booktabs} 

\usepackage{hyperref}

\usepackage{listings}
\usepackage{xcolor}

\usepackage{xcolor}
\usepackage{listings}
\lstdefinelanguage{PromptJSON}{
    basicstyle=\ttfamily\small,
    frame=single,
    backgroundcolor=\color{gray!5},
    breaklines=true,
    columns=fullflexible,
    numbers=none,
    showstringspaces=false,
    sensitive=true,
    morekeywords={system,user,question_},        
    alsoletter={[]},                   
    keywordstyle=\color{blue}\bfseries, 
    stringstyle=\color{orange!90!black}, 
}
\usepackage{fvextra}
\DefineVerbatimEnvironment{json}{Verbatim}{breaklines,fontsize=\small}
\usepackage{tcolorbox}
\tcbuselibrary{listings,breakable}
\usepackage{listings}
\usepackage{xcolor}
\usepackage{upquote} 
\usepackage{enumitem}
\usepackage{fvextra} 

\DefineVerbatimEnvironment{json}{Verbatim}{
  breaklines,
  breaksymbolleft={\tiny\ensuremath{\hookrightarrow}\,},
  fontsize=\small
}

\usepackage[preprint]{icml2026}

\usepackage{amsmath}
\usepackage{amssymb}
\usepackage{mathtools}
\usepackage{amsthm}

\usepackage[capitalize,noabbrev]{cleveref}

\theoremstyle{plain}

\theoremstyle{definition}

\theoremstyle{remark}

\usepackage[textsize=tiny]{todonotes}

\icmltitlerunning{Submission and Formatting Instructions for ICML 2026}

\begin{document}

\twocolumn[
  \icmltitle{GUI Knowledge Bench: Revealing the Knowledge Gap of VLMs in GUI Tasks}

  \icmlsetsymbol{equal}{*}

  \begin{icmlauthorlist}
  \icmlauthor{Chenrui Shi}{equal,bit,bigai}
  \icmlauthor{Zedong Yu}{equal,bupt,bigai}
  \icmlauthor{Zhi Gao}{equal,bit,bigai,msubit}
  \icmlauthor{Ruining Feng}{tsinghua,bigai}
  \icmlauthor{Enqi Liu}{bit,bigai}
  \icmlauthor{Yuwei Wu}{bit,msubit}
  \icmlauthor{Yunde Jia}{msubit}
  \icmlauthor{Liuyu Xiang}{bupt}
  \icmlauthor{Zhaofeng He}{bupt}
  \icmlauthor{Qing Li}{bigai}
\end{icmlauthorlist}

\icmlaffiliation{bit}{Beijing Institute of Technology, Beijing, China}
\icmlaffiliation{bigai}{State Key Laboratory of General Artificial Intelligence, BIGAI, Beijing, China}
\icmlaffiliation{bupt}{Beijing University of Posts and Telecommunications, Beijing, China}
\icmlaffiliation{tsinghua}{Tsinghua University, Beijing, China}
\icmlaffiliation{msubit}{Shenzhen MSU-BIT University, Shenzhen, China}

  \icmlcorrespondingauthor{Yuwei Wu}{yuwei.wu@msubit.edu.cn}
\icmlcorrespondingauthor{Qing Li}{qing.li@bigai.ai}

\centering \textbf{Project website}: \texttt{\href{https://computer-use-agents.github.io/GUI-Knowledge-Bench}{computer-use-agents.github.io/GUI-Knowledge-Bench}}
  \icmlkeywords{Machine Learning, ICML}

  \vskip 0.3in
]



\printAffiliationsAndNotice{}  
\begin{figure*}[]
    \centering
    \includegraphics[width=0.20\linewidth]{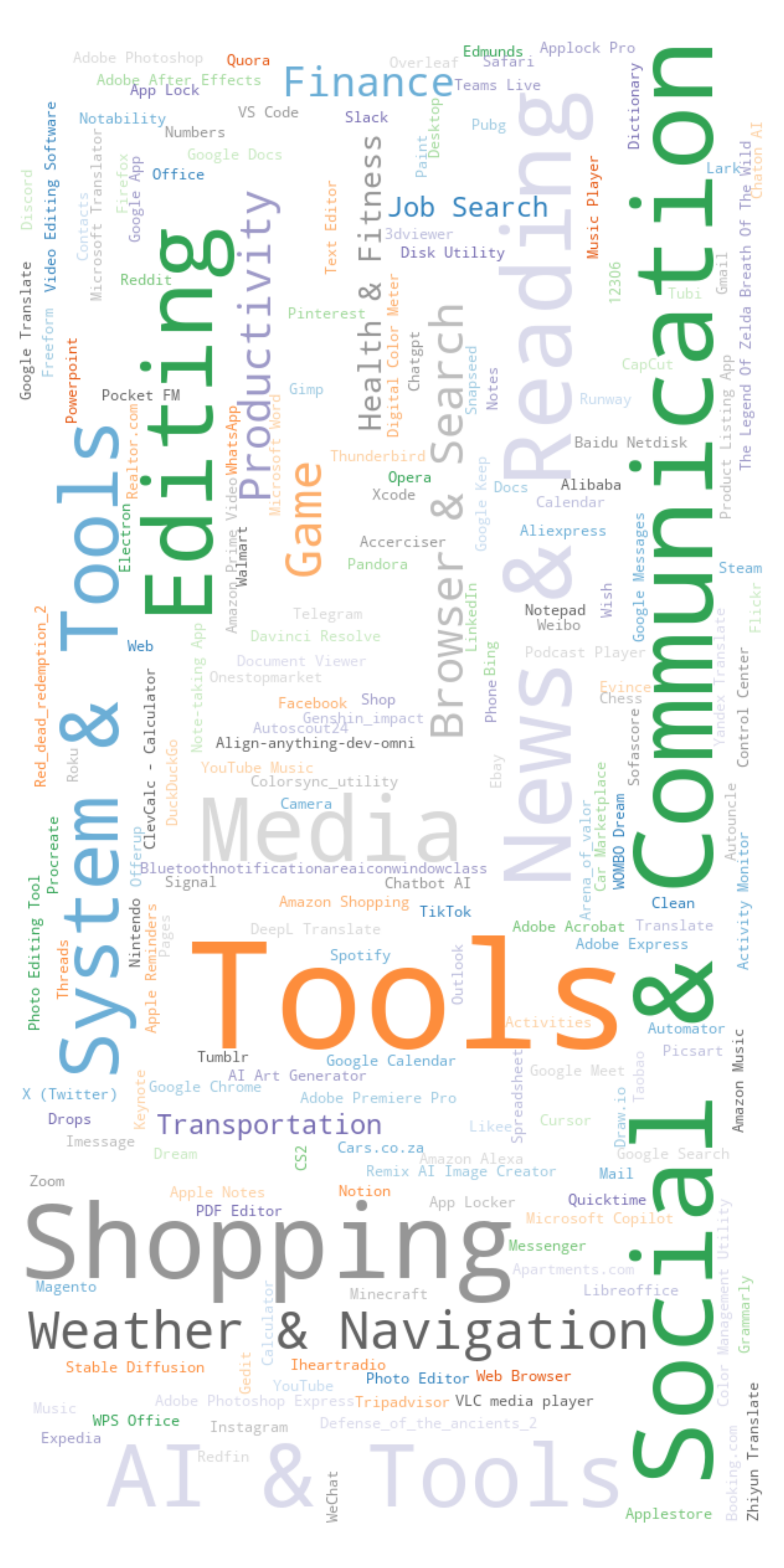}
    \includegraphics[width=0.75\linewidth]{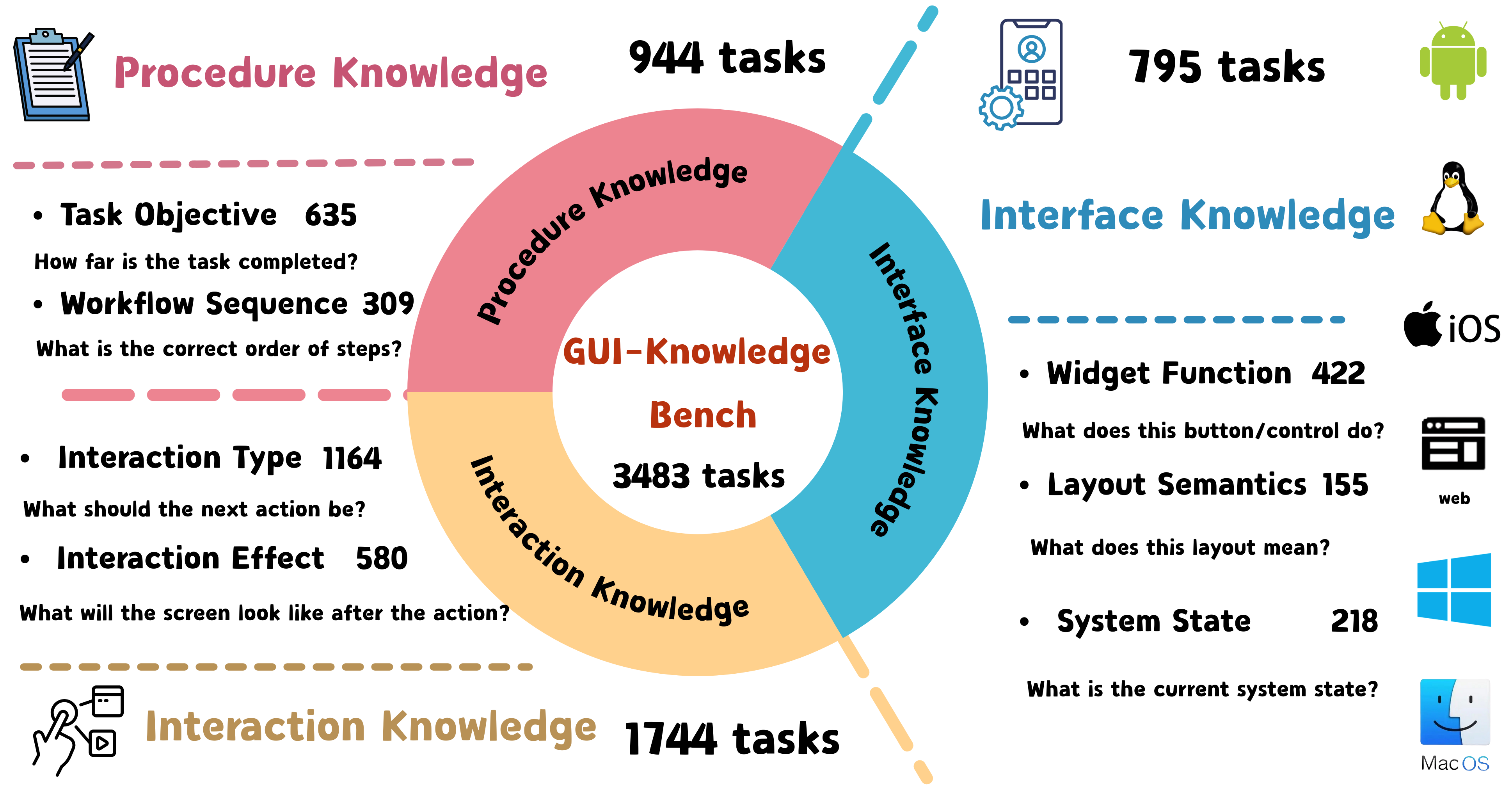}
    \caption{GUI Knowledge Bench: A benchmark evaluating VLMs on GUI knowledge across six platforms (Web, Android, MacOS, Windows, Linux, IOS). It measures three types of knowledge: Interface Knowledge, which evaluates understanding of GUI components, layout, and system state; Interaction Knowledge, which assesses the knowledge of GUI interaction conventions; and Procedure Knowledge , which tests whether a model knows the procedural knowledge of completing a GUI task.}
    \label{fig:banner}
\end{figure*}

\begin{abstract}
  Vision–language models (VLMs) have advanced graphical user interface (GUI) task automation but still lag behind humans. We hypothesize this gap stems from missing core GUI knowledge, which existing training schemes (such as supervised fine-tuning and reinforcement learning) alone cannot fully address. By analyzing common failure patterns in GUI task execution, we distill GUI knowledge into three dimensions: (1) interface knowledge about widget functions, layout semantics, and system states; (2) interaction knowledge about GUI interaction types and effects; and (3) procedure knowledge of task objectives and workflow sequences. We further introduce GUI Knowledge Bench, a benchmark with multiple-choice and yes/no questions across six platforms (Web, Android, MacOS, Windows, Linux, IOS) and 292 applications. Our evaluation indicates that current VLMs are generally aware of the functions of individual widgets, but lack the GUI-specific knowledge required to track system states, adhere to GUI interaction conventions, and assess task completion progress. Experiments on real-world GUI tasks further validate the close link between GUI knowledge and task success. By providing a structured framework for assessing GUI knowledge, our work supports the selection of VLMs with greater potential prior to downstream training and provides insights for building more capable GUI agents.

\end{abstract}

\input{section/intro}

\input{section/relatedwork}
\input{section/benchmark}
\input{section/result}
\newpage
\newpage

\bibliography{example_paper,reference_header}
\bibliographystyle{icml2026}

\newpage 

\newpage 

\section{Appendix}
\input{section/appendix}
\end{document}

%% file: section/intro.tex
\section{Introduction}

Graphical User Interface (GUI) task automation, such as booking a flight, editing a presentation, or configuring system settings, poses unique challenges for AI agents~\citep{wu2024oscopilotgeneralistcomputeragents,hong2024cogagent,xu2024androidlabtrainingsystematicbenchmarking,he2024webvoyager}. Recent approaches have leveraged vision–language models (VLMs) with techniques such as prompt engineering~\citep{agashe2025agent,xie2025scaling}, supervised fine-tuning (SFT)~\citep{wu2024atlas,hong2024cogagent,lin2025showui,liu2025infiguiagent, xu2024aguvis,zhang2025tonguiinternetscaletrajectoriesmultimodal}, and reinforcement learning (RL)~\citep{lian2025ui,luo2025gui,li2025efficient}, achieving strong task performance in many applications. 
However, GUI agents still fail in many real-world scenarios~\citep{xie2025osworld}.
In particular, agents may lack interface knowledge and procedure knowledge required to successfully accomplish a task. As illustrated in \cref{fig:MissingInterface&Instruction}, UITARS-1.5-7B fails to recognize the function of the view button for opening the notes section and does not know the correct procedure for adding notes. Moreover, agents may not adhere to correct GUI interaction conventions due to a lack of interaction knowledge, as shown in \cref{fig:real-InteractionPrediction}.

Our analysis suggests that a primary reason for these failures is that the used VLMs lack the necessary GUI knowledge. While prompt engineering, SFT, and RL can improve the reasoning, grounding, and planning abilities, they contribute little to injecting new GUI knowledge~\citep{ovadia2024finetuningretrievalcomparingknowledge},
which is essential for solving GUI tasks.
However, most existing benchmarks that primarily evaluate task success rate, focusing on the grounding~\citep{li2025screenspot,cheng2024seeclick,jurmu2008screenspot}, reasoning, and planning~\citep{lin2024videogui} capabilities of GUI models, and evaluating the GUI knowledge remains lacking.

In this paper, we introduce GUI-Knowledge Bench, a benchmark designed to systematically assess the extent of GUI knowledge encoded in VLMs, as shown in \cref{fig:banner}.
We categorize the GUI knowledge into three complementary aspects derived from common GUI task failure modes: (1) interface knowledge, which involves recognizing widget functions, layout semantics, and perceiving state information (e.g., enabled/disabled, selected/focused); (2) interaction knowledge, which involves assessing knowledge of GUI interaction conventions (e.g., how to interact with the interface elements, such as, toggle switch, a slide bar or a button); and (3) procedure knowledge, which evaluates the knowledge of workflow sequences for completing a GUI task. This categorization enables a systematic examination of which components of GUI knowledge are already present in current models and which remain underdeveloped.

To this end, our benchmark is deliberately designed to isolate GUI-specific knowledge from confounding factors.
We decouple the visual grounding ability by providing explicit visual markers in image, adopt multiple-choice and yes/no formats to reduce variability from open-ended generation, and remove samples that require the reasoning ability via human check.
These design enables a focused evaluation of whether models possess GUI knowledge, rather than their visual grounding, language generation, and abstract reasoning abilities. Along this way, we construct the benchmark from over 40,000 screenshots and 400 execution trajectories spanning 292 applications across six platforms (Web, Android, MacOS, Windows, Linux, IOS). Through a combination of automated generation and manual annotation, we obtain 3483 knowledge-centric questions that systematically evaluate VLMs' GUI knowledge. 

Our evaluation reveals that current VLMs are still short of enough knowledge in these three categories for completing real-world GUI tasks. 
First, although VLMs perform well at discerning functions of different widgets and understanding layout semantics, they have less knowledge for tracking system states.  
Second, VLMs underperform in GUI interaction conventions, showing difficulties in anticipating correct interaction effects and types. They frequently confuse click actions with other types of actions, a behavior commonly observed in many models. 
Third, VLMs struggle with the task objective. 
These findings highlight critical gaps in the internal GUI knowledge of current VLMs. 
Our contributions are as followed: 
\begin{itemize}
    \item On real-world GUI tasks, we reveal a close link between GUI knowledge and task success. Thus, we introduce GUI-Knowledge Bench to evaluate GUI knowledge in both general and GUI-tuned VLMs, across 292 applications on six platforms. 
    
    \item Our evaluation reveals key gaps in VLMs’ GUI knowledge, including system states, GUI interaction conventions, and task completion procedures, providing a knowledge-based guide for selecting potential models or improving GUI agents.
\end{itemize}

%% file: section/relatedwork.tex
\section{Related Work}

\subsection{GUI Agent} 
Progress in GUI task automation has largely relied on pretrained vision–language models (VLMs), with improvements driven by supervised fine-tuning (SFT), reinforcement learning (RL), and synthetic data generation. SFT-based methods train VLMs on large-scale GUI datasets to enhance element grounding and action prediction, as seen in OS-Atlas ~\citep{wu2024atlas}, CogAgent ~\citep{hong2024cogagent}, and ShowUI~\citep{lin2025showui}, while multi-stage pipelines such as InfiGUIAgent~\citep{liu2025infiguiagent} and Aguvis~\citep{xu2024aguvis} further inject reasoning and planning abilities with synthetic data. RL approaches, including UI-AGILE~\citep{lian2025ui} and GUI-R1~\citep{luo2025gui}, refine action selection through long-horizon rewards or policy optimization, sometimes achieving superior performance with less training data. To address data scarcity, OS-Genesis~\citep{sun2025genesis} and UI-Genie~\citep{sun2024genesis} generate high-quality synthetic trajectories, while multi-agent systems such as GUI-OWL and  Mobile-Agent-v3~\citep{wanyan2025look} decompose perception, reasoning, and planning across modules to improve robustness in long-horizon tasks.

Existing approaches focus on improving GUI execution strategies via imitation of expert trajectories, reward shaping, or modular design, without fundamentally enriching the model’s internal GUI knowledge. 
The obtained agents may still fall short in unfamiliar applications or complex system states. 
To address this gap, our work systematically evaluates these foundational knowledge deficiencies and introduces a benchmark that identifies missing GUI knowledge in VLMs.

\subsection{GUI Benchmark}
Existing benchmarks generally fall into three categories. 
Action-level benchmarks focus on the precision of low-level operations such as mouse and keyboard inputs, and accurate element grounding. Examples include ScreenSpot-Pro~\citep{li2025screenspot} for grounding challenges in professional high-resolution interfaces, and SeeClick~\citep{cheng2024seeclick} and ScreenSpot~\citep{jurmu2008screenspot} for cross-environment grounding. 
Plan-level evaluations extend beyond single actions to hierarchical execution. VideoGUI~\citep{lin2024videogui}, for instance, evaluates GUI agents with high-level and mid-level planning. 
Task-level benchmarks emphasize real task success in simulated environments, such as OSWorld~\citep{xie2025osworld}, OSWorld-Verified~\citep{osworld_verified}, MacOSworld~\citep{yang2025macosworld}, and AndroidWorld ~\citep{rawles2024androidworld}.

In contrast, our benchmark is explicitly designed to evaluate whether a model possesses sufficient GUI knowledge, independent of grounding, planning, and reasoning, as shown in \cref{tab:benchmark_comparison}. 
To this end, we decouple element grounding from our evaluation by providing visual markers on screenshots, thereby minimizing grounding-related confounds. All questions are formulated as simple multiple-choice or yes/no queries that directly evaluate GUI knowledge, rather than requiring multi-step reasoning or text generation.
Our benchmark carefully categorizes the GUI knowledge into three complementary aspects derived from common failure modes in GUI task automation, \textit{i.e.}, interface knowledge, interaction knowledge, and procedure knowledge. 

A few recent efforts also involve GUI knowledge, such as MMBench-GUI~\citep{wang2025mmbench}, which tests content understanding and widget semantics, and Web-CogBench~\citep{guo2025web}, which evaluates cognitive reasoning in web navigation. Different from the two benchmarks that remain narrow in application scopes and domain knowledge coverage, our benchmark offers a systematic and comprehensive evaluation of GUI knowledge, spanning multiple platforms and applications, thereby providing a more complete evaluation of VLM's GUI knowledge.

\begin{table}[t]
\caption{Comparison of existing GUI benchmarks and our benchmark across evaluation scope, operating system coverage, application diversity, and data scale. Our benchmark systematically spans multiple OS and applications with a comprehensive scope of GUI knowledge evaluation.}
\centering
\resizebox{\linewidth}{!}{
\begin{tabular}{p{6cm} p{1.5cm} p{0.5cm} p{1cm} p{2cm}}
\hline
\textbf{Benchmark} & \textbf{Scope} & \textbf{OS} & \textbf{Apps} & \textbf{Task Num.} \\
\hline
ScreenSpot-Pro~\citep{li2025screenspot}  & Action & 3 & 23 & 1581 \\
SeeClick~\citep{cheng2024seeclick} & Action & 5 & 20+ & 1272 \\
VideoGUI~\citep{lin2024videogui}  &Task  & 1 & 11 & 463 \\
OSWorld~\citep{xie2025osworld}  &Task  & 1 & 9 & 369 \\
MacOSworld~\citep{yang2025macosworld} &Task  & 1 & 30 & 202 \\
AndroidWorld~\citep{rawles2024androidworld}  &Task  & 1 & 20 & 116 \\
MMBench-GUI~\citep{wang2025mmbench}  &Knowledge  & 6 & 98 & 8000+ \\
Web-CogBench~\citep{guo2025web} &Knowledge  & 1 & 14  & 876 \\
\hline
\textbf{GUI-Knowledge-Bench}  &Knowledge  & \textbf{6} & \textbf{292} & \textbf{3483} \\
\hline
\end{tabular}}

\label{tab:benchmark_comparison}
\end{table}

\begin{figure*}[t]
    \centering
    \includegraphics[width=1\linewidth]{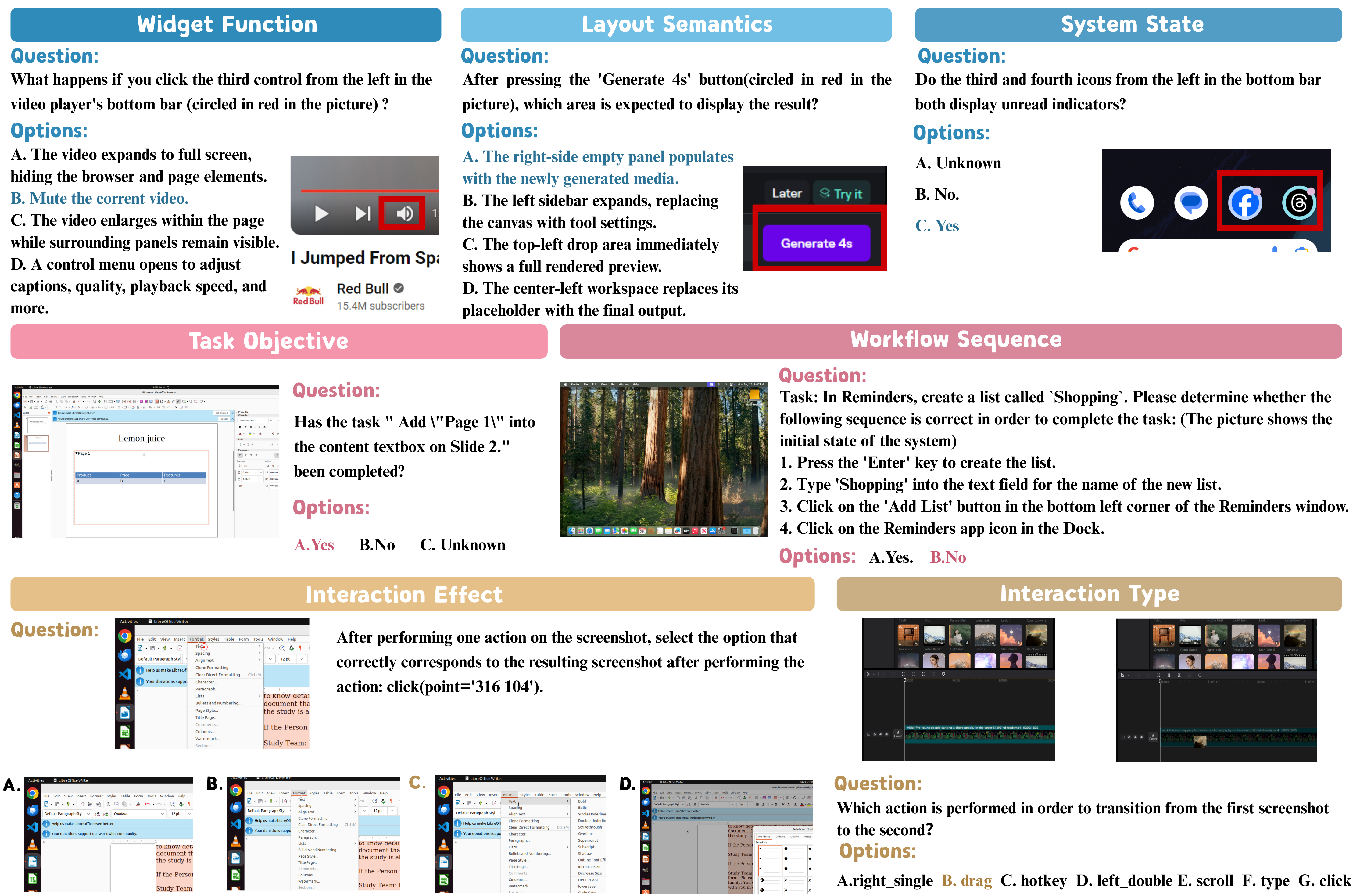}
    \caption{Example questions of GUI Knowledge Bench. }
    \label{fig:Question Examples}
    \vspace{-3mm}
\end{figure*}

%% file: section/benchmark.tex
\section{GUI Knowledge Bench}

\subsection{Benchmark Overview}
We introduce GUI Knowledge Bench, and some examples are shown in~\cref{fig:Question Examples}, a benchmark for systematically evaluating the knowledge required by VLMs to complete GUI tasks. 
Based on common failure patterns observed in GUI task execution, we identify three complementary dimensions:
(1) \textbf{Interface Knowledge}, which examines whether models know widget functions, layout semantics, and system states in GUI screenshots;
(2) \textbf{Interaction knowledge}, which examines whether models know the types and effects of common GUI interactions; and
(3) \textbf{Procedure knowledge}, which assesses whether models know the task objective and the workflow sequence for completing GUI tasks.
Together, these dimensions capture the core knowledge required for reliable GUI task completion and form the foundation of our benchmark.

\subsection{Benchmark Design}
Our benchmark is deliberately designed to isolate GUI-specific knowledge from confounding factors (e.g., grounding, planning, and reasoning). 
To this end, we decouple visual grounding by providing explicit visual markers. For example, questions directly refer to widgets enclosed by red bounding boxes (visual markers), asking for their functionality. 
We further adopt multiple-choice and yes/no question formats to reduce variability from open-ended text generation, and avoid reasoning-intensive samples. For instance, queries are phrased to require a single-step factual judgment about the current interface state, rather than multi-hop or compositional reasoning.
These choices enable a stable and focused evaluation that targets whether models have interface knowledge, interaction knowledge and procedure knowledge, rather than their language generation fluency or abstract reasoning ability. By constraining the answer space and grounding each question in a single interface state or interaction, model performance more directly reflects GUI knowledge and interaction conventions. We further validate the robustness of this evaluation protocol through additional ablation studies, including option-order randomization and free-form answer evaluation.
Results show that model performance remains consistent under option-order randomization and free-form answer evaluation, indicating that the observed evaluation results are not driven by superficial multiple-choice cues or linguistic biases. Instead, the benchmark reliably measures models’ GUI knowledge demonstrating the stability and validity of our evaluation protocol. Please refer to appendix \cref{EvaluationProtocol} for further details. 


\subsection{Data Sources and Collection Pipeline}
To build GUI Knowledge Bench, we aggregate data from multiple sources to ensure both trajectory-level interaction coverage and diverse standalone screenshots.

We leverage existing benchmarks such as GUI-Odyssey~\citep{lu2024gui} and VideoGUI~\citep{lin2024videogui}, which provide screenshots paired with tasks and action annotations. In addition, we collect new trajectories by running UI-Tars-7B agents in environments including OSWorld and MacOSWorld, capturing realistic interaction sequences across both mobile and desktop platforms.

To increase visual diversity and cover a wider range of application interfaces and operating systems, we further gather standalone GUI screenshots. Specifically, we sample from ScreenSpot v2 and extract representative key frames from YouTube tutorials, ensuring coverage of real-world applications, operating systems, and interface layouts. For less common interactions, we manually perform operations on MacOS, Linux, and Windows, recording screenshots and corresponding actions.

Together, these sources yield a heterogeneous pool of GUI images and trajectories. From this pool, we construct task-specific question–answer pairs for each evaluation dimension, ensuring sufficient diversity and coverage while minimizing redundancy.
 Please refer the appendix \cref{datasetstatistics}
for detailed statistics of our benchmark.

\subsection{Interface Knowledge}

Interface knowledge encloses three aspects: (i) widget function, i.e., recognizing the functions of common interface elements (e.g., three vertical dots for settings, speech bubbles for messaging apps); (ii) system states, such as knowing whether a button is enabled/disabled, whether a paragraph is selected/focused, or whether a switch is toggled on/off; and (iii) layout semantics, where spatial arrangement in GUI encodes critical information (e.g., distinguishing departure and arrival cities by their relative positions, identifying senders and receivers in an email, or understanding file hierarchy from indentation). 

\textbf{Task Definition.}  
We formalize the evaluation as a unified multiple-choice question-answering task. Given a question $q$, a set of candidate options $O$, and a screenshot $S$, the model is required to select the correct answer $o^{*}$ and provide its thought $t$:
$\text{VLM}: (S, q, O) \mapsto (t, o^{*}).$

To reduce the burden of visual grounding, we explicitly mark the relevant interface elements in the screenshot $S$ using visual markers such as bounding boxes or red dots. 
When referring to these elements in the question text, we directly specify the highlighted region (e.g., ``the widget enclosed by the bounding box'' or ``the area indicated by the red dot''), thereby eliminating the need for implicit grounding. 
This design ensures that the evaluation focuses on whether the model possesses the required GUI knowledge rather than its visual grounding ability.

\textbf{Task Collection and Curation.}
To construct the evaluation set, we first have human annotators design an initial set of seed questions based on the collected GUI screenshots. We then leverage GPT-5 to expand this pool with additional candidate questions, increasing diversity while maintaining relevance. Questions that can be answered based solely on the text, without viewing the screenshot, are removed using Qwen-2.5-VL-7B to ensure the necessity and relevance to the screenshots. Finally, the remaining questions are manually verified for correctness, and relevant regions in the screenshots are annotated to support precise visual grounding. This pipeline ensures that the evaluation focuses on interface knowledge knowledge rather than being confounded by grounding or reasoning.

\subsection{Interaction Knowledge}

GUI interactions follow symbolic and platform-specific rules (e.g., toggling a switch with `click' actions and opening a file with `double click' actions), which are often subtle and context-dependent. Without a proper understanding of these interaction conventions, models cannot reliably predict the effect of GUI interactions or correctly interact with GUIs to complete a task. 

Interaction knowledge is evaluated through two complementary tasks:  
(i) Interaction Effect. The model is provided with a screenshot $S$ in which a candidate interaction (e.g., \textit{click}) is explicitly annotated by visual markers indicating its execution location, as shown in \cref{fig:Question Examples}.
The model is required to select the resulting screenshot $S'$ from a set of candidates, evaluating whether it understands the effect of executing the specified interaction at the given location.

(ii) Interaction Type. 
The model is given two consecutive screenshots $(S, S')$ and must infer the interaction that caused the transition. Candidate options correspond to different interaction arguments, with their execution locations explicitly annotated in the screenshot.
The model selects among these candidates without the need to generate precise coordinates.
Importantly, in our evaluation, models are not required to predict or localize interaction parameters.
All candidate interactions are fully specified through visual annotations, ensuring that the evaluation does not assess visual grounding.
Instead, it isolates interaction knowledge—namely, understanding which interaction types are valid in a given GUI context and what parameters are required to produce a given state transition.



\textbf{Task Definition.}  
We formalize GUI interaction dynamics as a state–action transition $S + a \rightarrow S'$, where $S$ represents the current screenshot, $S'$ the consequent screenshots and $a$ the action $a = (a_\text{type}, a_\text{param})$.  
(i) Interaction Effect.
The model is given $S$ and $a$, and is required to select the resulting screenshots from a set of candidate screenshot options $O$: $ \text{VLM}: (S, a, O) \mapsto S'$.
(ii) Interaction Type.
The model is given two consecutive screenshots $(S, S')$ and a set of candidate interaction types $O_\text{type}$, and must select the correct interaction type $a_\text{type}$:  $\text{VLM}: (S, S', O_\text{type}) \mapsto a_\text{type}$.
Given the correct interaction type $a_\text{type}$ and the same state pair $(S, S')$, the model selects the correct interaction parameters from a candidate set $O_\text{param}$:  $\text{VLM}: (S, S', a_\text{type}, O_\text{param}) \mapsto a_\text{param}$.

\textbf{Task Collection and Curation.} 
For interaction effect, candidate screenshots include the preceding screenshots, the true next screenshot, and visually similar but different screenshots sampled from current trajectories. For interaction type, the model must choose from the full set of possible actions defined for the platform, with a unified action space across different platforms. (seven actions for desktop platforms and four actions for mobile platforms) For interaction parameter, 
for clicks, bounding boxes of candidate elements are identified using OmniParser, and nearby but incorrect coordinates are sampled as options; for drags, distractors include reversed directions, shortened distances, or swapped start and end points; for scrolls, distractors vary in direction (up, down, left, right); for typing, inputs are perturbed with case changes, partial deletions, or common typos; and for hotkeys, distractors are drawn from a predefined set of common shortcuts. 

\subsection{Procedure Knowledge}

Procedure knowledge evaluates whether a model can track GUI task completion progress and have the procedural knowledge of completing a GUI task.  
The questions do not ask the model to construct a plan or generate sub-steps. Instead, the complete operation sequence is already provided in these questions to minimize the reasoning load and text generation burden. 
We assess two complementary abilities: (i) task objective, which evaluates whether a model can determine if a task has been successfully completed based on history screenshots; and (ii) workflow sequence, which evaluates whether a model knows the right procedure to complete a task. 
Questions in our benchmark are designed that once the procedure knowledge is known, the answer becomes immediately obvious.

\textbf{Task Definition.}  
For task objective, the model receives a natural-language task description $t$ and history screenshots, and must select the correct option $o^* \in \{\text{Yes}, \text{No}, \text{Unknown}\}$ indicating whether the task is completed:  
$\text{VLM}: (S_{1:T}, t, O) \mapsto o^*$.  
For workflow sequence, the model is given a natural-language task description $t$, current screenshot $S$ and a set of candidate orderings $O = \{\pi_1, \pi_2, \dots, \pi_m\}$, where each $\pi_i$ is a possible permutation of the operation steps. The model must select the correct ordering $\pi^{*}$ from $O$:  
$\text{VLM}: (t, S, O) \mapsto \pi^{*}$.

\textbf{Task Collection and Curation. }
For task objective, human annotators label each trajectory as successful or unsuccessful based on last one to five screenshots of the trajectory, and some successful trajectories are augmented by removing the final screenshot to create unsuccessful ones. For workflow sequence, operation plans are first generated by Chat-GPT-5 and then verified by annotators. The annotated steps are automatically shuffled to form multiple-choice ordering questions, with longer sequences retaining initial steps and only permuting later steps. For shorter sequences, additional question formats are created by converting the shuffled sequence into Yes/No/Unknown questions, or into operation-level fill-in-the-blank questions with distractor steps. Tasks solvable without observing screenshots are filtered out using Qwen-VL-2.5-7B ensuring the relevance to screenshots.




%% file: section/result.tex
\section{Benchmarking VLMs }
\begin{figure*}[t]
    \centering
    \begin{subfigure}[t]{0.48\linewidth}
        \centering
        \includegraphics[width=\linewidth]{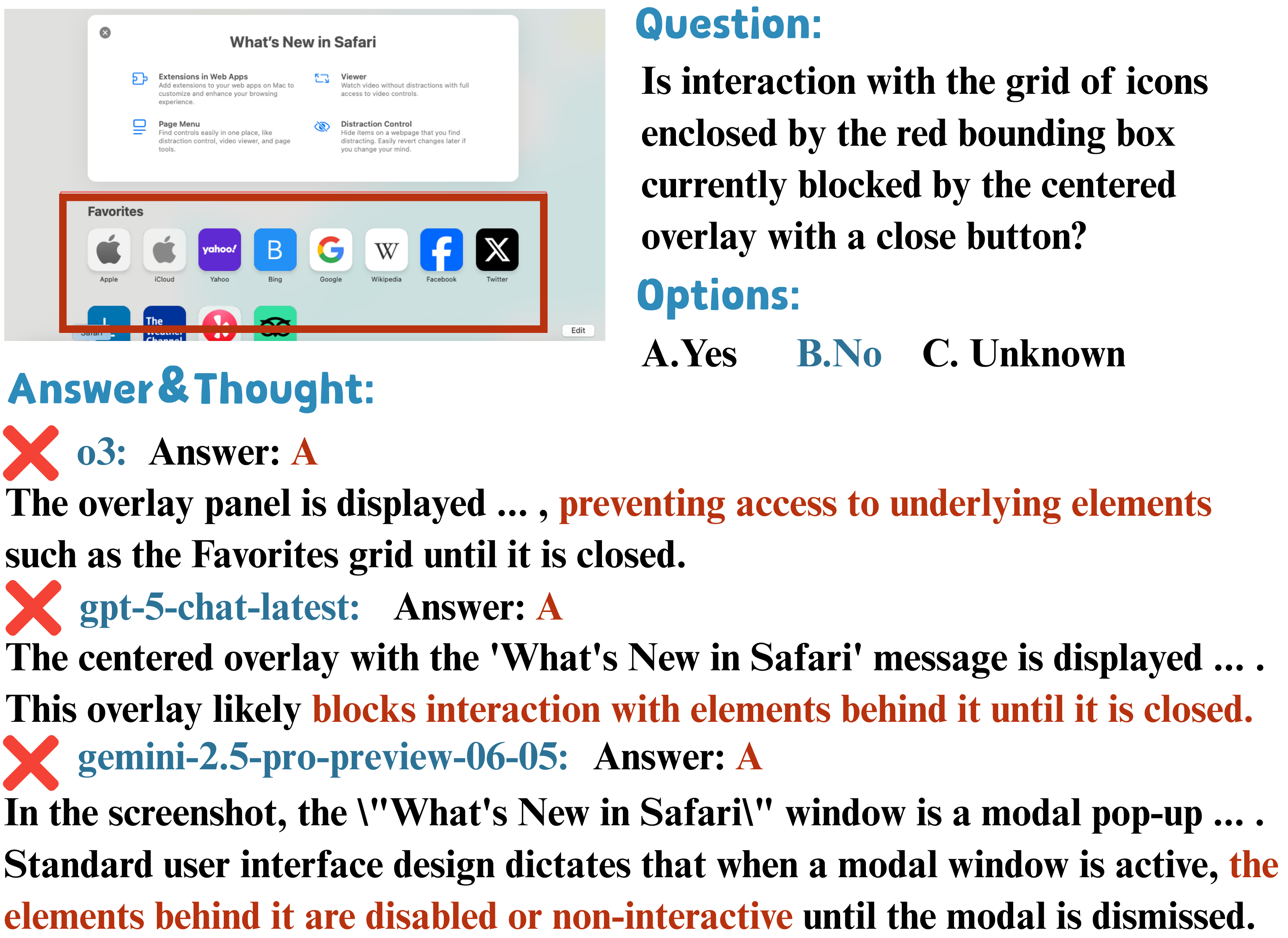}
        \caption{Failure cases of interface knowledge questions.}
        \label{fig:interface-perception-errors}
    \end{subfigure}
    \hfill
    \begin{subfigure}[t]{0.48\linewidth}
        \centering
        \includegraphics[width=\linewidth]{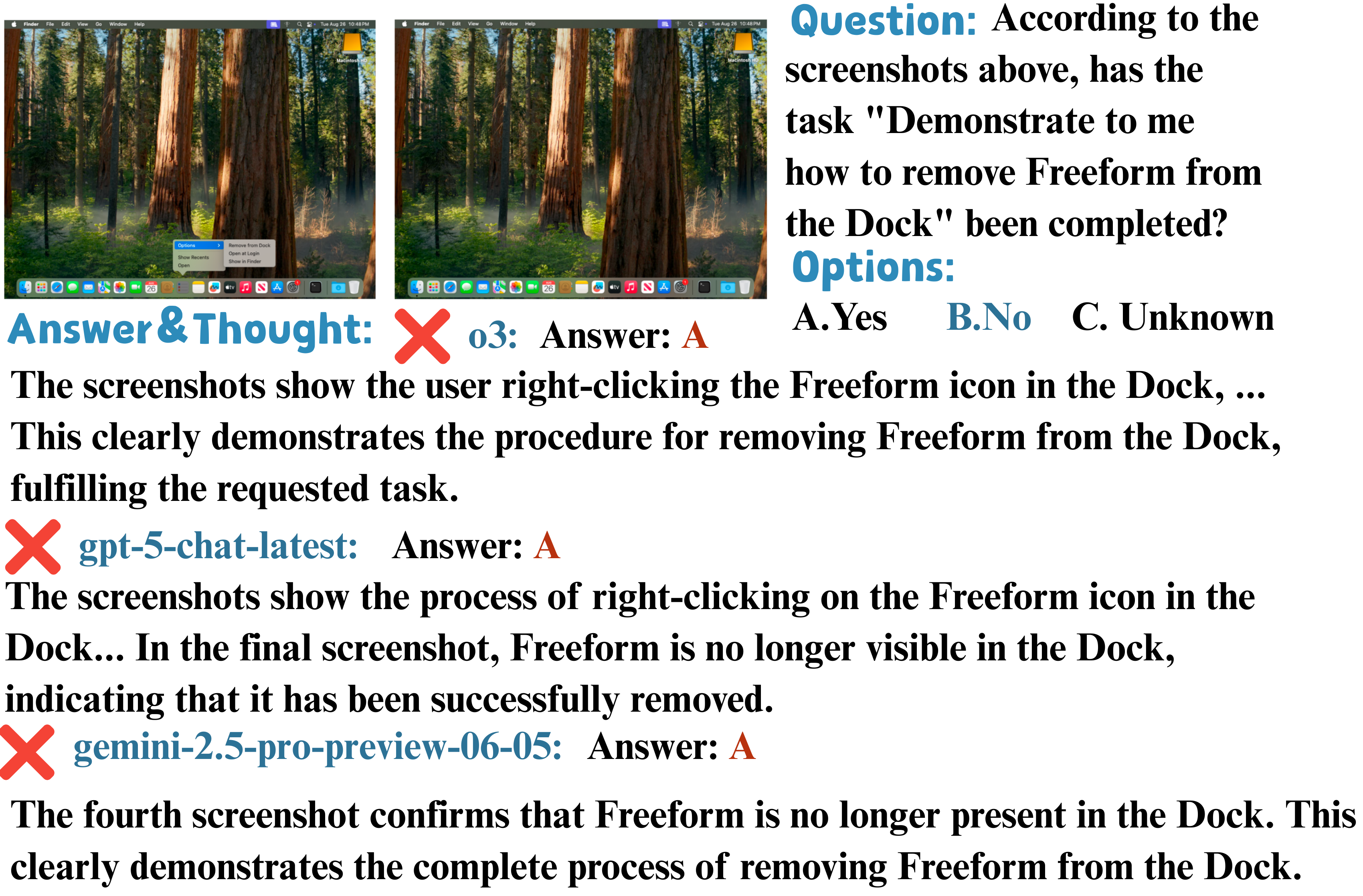}
        \caption{Failure cases of procedure knowledge questions.}
        \label{fig:instruction-understanding-errors}
    \end{subfigure}

    \caption{Representative failure cases in GUI Knowledge Bench.}
    \label{fig:failure-cases-gui}
\end{figure*}


\begin{table*}[ht]
\centering
\caption{Performance on GUI Knowledge Bench across interface knowledge, interaction knowledge and procedure knowledge. Bold numbers indicate the best results in each sub-task.}
\resizebox{\textwidth}{!}{%
\begin{tabular}{c|ccc|ccc|cc|c}
\hline
\small Model & \multicolumn{3}{c|}{\small Interface Knowledge} & \multicolumn{3}{c|}{\small Interaction Knowledge} & \multicolumn{2}{c|}{\small  Procedure Knowledge} & \multicolumn{1}{c}{\small Overall} \\
 & \small state & \small widget & \small layout & \small effect & \small type & \small parameter & \small objective & \small workflow &  \\
\hline
{\small O3~\citep{openai2025o3systemcard} } & \textbf{83.03\%} & 84.12\% & \textbf{88.39\%} & \textbf{74.83\%} & \textbf{75.98}\% & 45.75\% & 69.45\% & \textbf{95.47\%} & \textbf{73.30\%} \\
{\small Gemini-2.5-Pro~\citep{comanici2025gemini25pushingfrontier} } & 81.19\% & \textbf{84.36}\% & 87.10\% & 71.03\% & 73.25\% & \textbf{46.97\%} & 67.72\% & 92.56\% & 71.69\% \\
{\small GPT-5-Chat~\citep{openai2025gpt5systemcard} } & 78.90\% & 84.12\% & \textbf{88.39\%} & 71.55\% & 71.55\% & 43.85\% & 68.98\% & 91.26\% & 70.97\% \\
{\small Qwen3-vl-8b-thinking~\citep{qwen3technicalreport} } & 68.81\% & 76.30\% & 83.23\% & 67.07\% & 70.36\% & 40.73\% & 64.09\% & 91.26\% & 66.81\% \\
{\small Claude-Sonnet-4-5~\citep{anthropic2025claude45} } & 74.77\% & 81.52\% & 82.58\% & 49.83\% & 70.19\% & 43.33\% & \textbf{70.30\%} & 91.56\% & 66.53\% \\
{\small Qwen2.5VL-72B~\citep{bai2025qwen25vltechnicalreport} } & 69.27\% & 77.49\% & 80.00\% & 61.72\% & 64.91\% & 38.99\% & 62.20\% & 85.44\% & 63.88\% \\
{\small Doubao-V-Pro~\citep{guo2025seed15vltechnicalreport}  } & 72.48\% & 83.65\% & 81.29\% & 67.24\% & 75.64\% & 41.07\% & 33.07\% & 94.17\% & 63.42\% \\
{\small Claude-Sonnet-4~\citep{anthropic2025claude4systemcard}  } & 70.18\% & 78.44\% & 78.06\% & 41.90\% & 62.52\% & 42.11\% & 65.20\% & 94.82\% & 62.16\% \\
{\small Qwen2.5VL-7B~\citep{bai2025qwen25vltechnicalreport} } & 53.21\% & 67.77\% & 60.00\% & 51.72\% & 50.60\% & 39.34\% & 16.22\% & 48.87\% & 45.16\% \\
{\small UITARS-1.5-7B~\citep{qin2025uitarspioneeringautomatedgui} } & 49.54\% & 59.48\% & 59.35\% & 22.24\% & 59.11\% & 34.32\% & 38.74\% & 55.34\% & 44.27\% \\
{\small GUI-OWL-7b~\citep{ye2025mobileagentv3fundamentalagentsgui} } & 60.09\% & 64.93\% & 63.23\% & 21.55\% & 55.37\% & 36.05\% & 21.26\% & 39.81\% & 40.74\% \\
{\small GLM-4.5~\citep{5team2025glm45agenticreasoningcoding} } & 49.54\% & 48.10\% & 53.55\% & 27.07\% & 17.55\% & 35.53\% & 28.98\% & 91.91\% & 38.10\% \\
\hline
\end{tabular}}
\label{tab:OverAll}
\end{table*}

\subsection{Settings}
We evaluate multiple open-source and closed-source models on the GUI Knowledge Bench. 
The closed-source set includes Claude-Sonnet-4-5~\citep{anthropic2025claude45}, Claude-Sonnet-4~\citep{anthropic2025claude4systemcard}, Doubao-V-Pro (Doubao-1.5-Thinking-Vision-Pro-250428)~\citep{guo2025seed15vltechnicalreport}, Gemini-2.5-Pro~\citep{comanici2025gemini25pushingfrontier}, GPT-5-chat~\citep{openai2025gpt5systemcard}, O3\citep{openai2025o3systemcard}, and GLM-4.5~\citep{5team2025glm45agenticreasoningcoding}. 
The open-source set covers Qwen2.5-72B (Qwen2.5-VL-72B-Instruct), Qwen2.5-7B (Qwen2.5-VL-7B-Instruct)~\citep{bai2025qwen25vltechnicalreport}, Qwen3-vl-8b-thinking~\cite{qwen3technicalreport}. Besides we also include GUI finetune models such as UITARS-1.5-7B~\citep{qin2025uitarspioneeringautomatedgui} and GUI-OWL-7b~\citep{ye2025mobileagentv3fundamentalagentsgui}. 
Apart from necessary model-specific settings, all other parameters (e.g., temperature, top-p) were kept consistent across evaluations. Please refer the appendix \cref{messagetemplate} for the detailed message template for each knowledge categories. 



\subsection{Benchmarking Results}
Table~\ref{tab:OverAll} summarizes the performance of all evaluated models on the three knowledge categories. Experimental results highlight the following key observations. 

First, o3 achieves strong performance across multiple metrics, consistent with its high success rate in real GUI tasks. In the OSWorld benchmark under the Agent framework category, many competitive agents leverage o3 (e.g., Agent-S2.5 w/ O3 50-step version and 100-step version, Jedi-7B w/ O3 w/ 50-step version and 100-step version).~\citep{agashe2025agent,xie2025scaling}
Second, 
UITARS-1.5-7B, trained on Qwen2.5VL-7B, improves procedure knowledge but exhibits a decline in interface knowledge. 
We attribute this to post-training adaptations for GUI task execution, which bias the model toward end-to-end task completion and reduce its instruction-following robustness on structured QA prompts. Notably, UITARS-2\cite{wang2025ui} incorporates continual pre-training to broaden general GUI knowledge. 
This design suggests an explicit attempt to compensate for GUI knowledge degradation introduced during task-focused post-training. Third, smaller models retain limited knowledge, suggesting that retrieval-augmented generation or knowledge-base integration may be a viable approach to enhance GUI agent performance.

\subsection{Error Analysis and Discussion}
\textbf{Interface Knowledge.} 
Most models handle widget functions and layout semantics well, but struggle with system state prediction. As shown in \cref{fig:interface-perception-errors}, in Safari, an update notification is mistaken for a blocking pop-up, leading to incorrect answers.

\textbf{Interaction Knowledge.}
On desktop, models often confuse click, double-click, and right-click in different situations. This is partly because different operating systems and applications treat these interactions differently: in some contexts, single, double, or right clicks can substitute for each other, while in others the distinction is strict. 
As a result, less frequent actions like double-click or right-click are more prone to misprediction, especially for smaller models. Please refer to appendix \cref{actionconfusion} to see the confusion matrix of action type prediction.

\textbf{Procedure Knowledge.}
Our benchmark highlights a failure in task objective. In Figure \ref{fig:instruction-understanding-errors}, three models (o3, gpt-5-chat-latest, and gemini-2.5-pro) all respond “Yes,” claiming that the Freeform icon is gone from the Dock, even though a different app was removed. This indicates that the models fail to correctly track the task objective, leading to incorrect conclusions about task completion.


%

\subsection{GUI Knowledge for Real-world GUI Tasks}
This section examines the role of three types of knowledge in enabling successful real-world GUI task execution. 

\subsubsection{Qualitative Analysis}
\begin{figure}[h]
    \includegraphics[width=\linewidth]{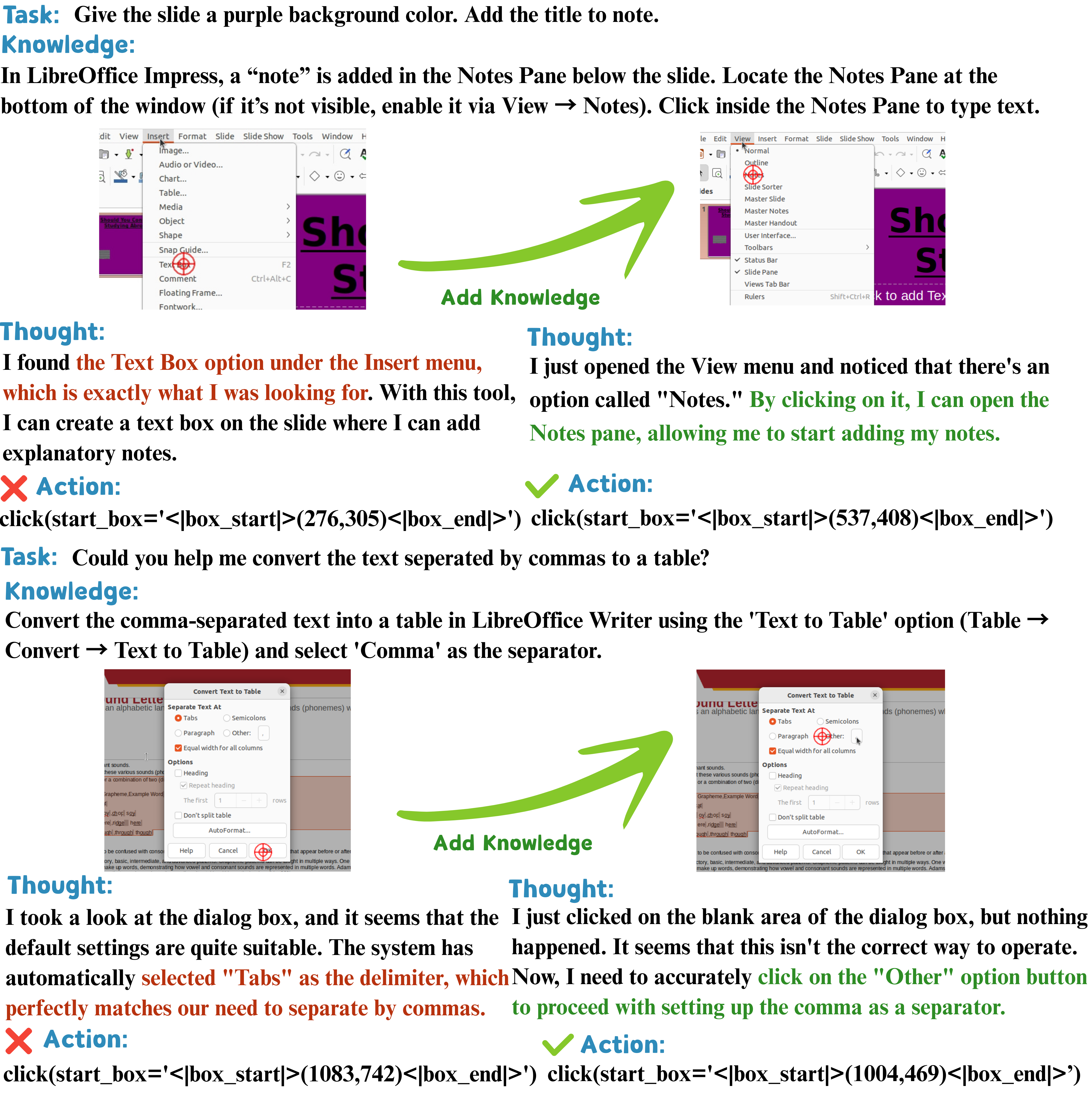}
    \caption{Effect of missing interface knowledge and procedure knowledge.}
    \label{fig:MissingInterface&Instruction}
\end{figure}
\textbf{Effect of Missing Interface Knowledge and Procedure Knowledge.} 
In our evaluation, we identify several tasks from OSWorld that UITARS-1.5-7B consistently fails to solve even under the pass@32 setting. We attribute these failures to missing interface and procedure knowledge required for completing GUI tasks.
Representative failure cases are shown in Figure~\ref{fig:MissingInterface&Instruction}.
In the task of adding a note, the model repeatedly inserts comments or text boxes, reflecting missing interface knowledge of the \textit{View} button for enabling the Notes pane, as well as missing procedure knowledge of the correct workflow (enabling the pane before adding content). Similarly, in the task of converting comma-separated text into a table, the model fails to specify the delimiter, indicating a lack of interface knowledge about the delimiter option and procedure knowledge that selecting delimiter opiton is a required step.
These cases indicate that the observed failures stem from the absence of application-specific interface and procedure knowledge.
Importantly, once the required knowledge is explicitly provided, the model is able to complete these tasks successfully.

\begin{table}[h]
\centering
\caption{Effect of injecting procedure knowledge on UITARS-1.5-7B.}
\begin{tabular}{l c}
\hline
Method & Pass@1 (\%) \\
\hline
UITARS-1.5-7B (base) & 24.81 \\
+ GPT-4o plan & 27.59 \\
+ OSWorld-human plan & 28.20 \\
+ o3 plan & \textbf{30.79} \\
\hline
\end{tabular}
\vspace{-4mm}
\label{tab:plan-augmentation}
\end{table}

\begin{figure}[t]
    \includegraphics[width=\linewidth]{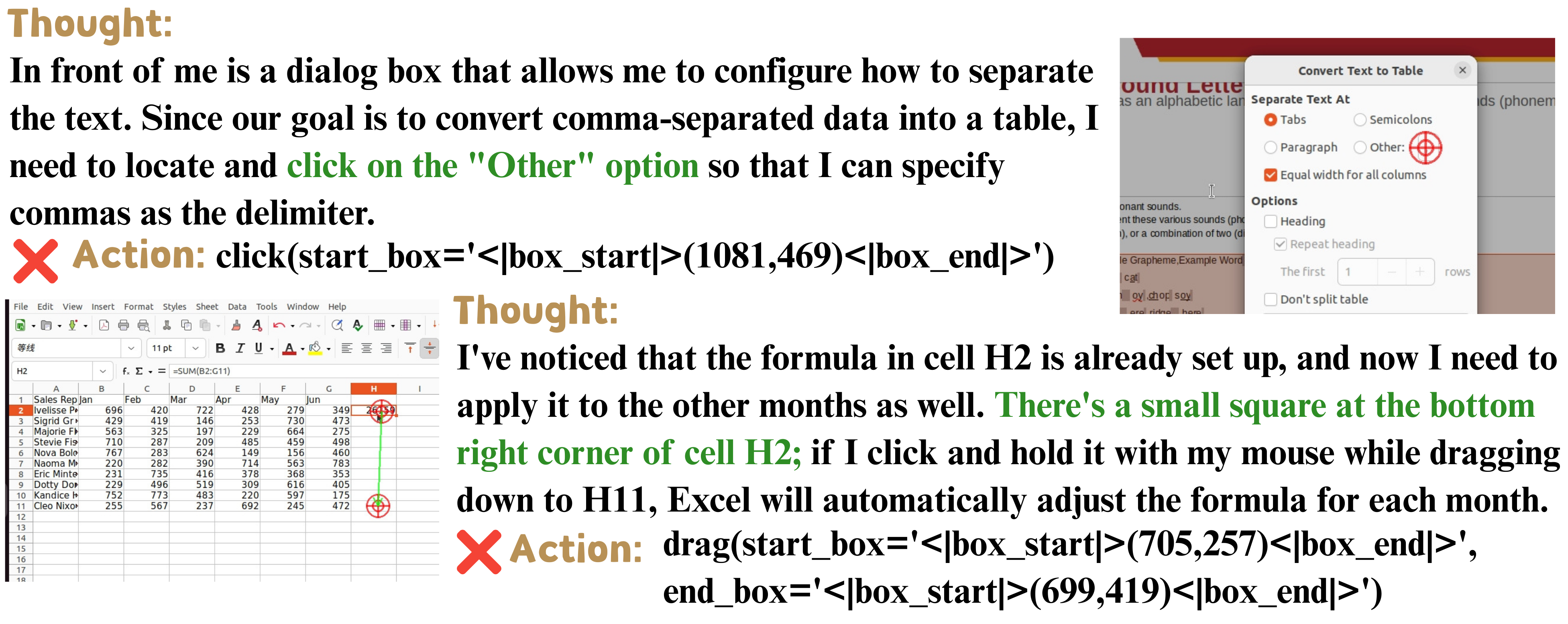}
    \caption{Effect of missing interaction  knowledge.}
    \label{fig:real-InteractionPrediction}
\end{figure}

\textbf{Effect of Missing Interaction Knowledge.}
As shown in Figure \ref{fig:real-InteractionPrediction}, many errors of models occur because it lacks knowledge for localizing interface elements correctly. 
Prior work has improved localization using masks, accessibility trees, or APIs. Another promising approach is to leverage actions themselves for self-verification, using visual prompts to check if the executed action was correct. 

\subsubsection{Quantitative Analysis}
\textbf{The Impact of Procedure Knowledge.}
We generated knowledge about operation plans wiht Chat-GPT-4o and o3 conditioned on task instructions, and used human-annotated operation plans from OSWorld-human. Each plan was appended to the original task description to inject procedure knowledge for UITARS-1.5-7B. Results are summarized in Table~\ref{tab:plan-augmentation}.
These results show that providing knowledge about operation plans improves task performance, highlighting the importance of procedure knowledge for task completion. Notably, o3-generated plans achieve the largest gain, surpassing human-annotated plans and aligning with o3's top performance across our benchmark evaluations.

\textbf{The Impact of Interface and Interaction knowledge.}
While the previous experiment demonstrates the benefit of injecting procedure knowledge, interface and interaction knowledge are intrinsically difficult to inject externally. Procedure knowledge can be provided as explicit step-by-step guidance, but interface and interaction knowledge—such as understanding a button’s function or the correct way to manipulate a widget—is often implicit and task-dependent, and cannot be fully identified prior to execution.
To address this, we conduct a validation study that mirrors the the procedure knowledge analysis for interface knowledge and interaction knowledge. Concretely, we transform 39 questions in our benchmark into practical GUI tasks, where the knowledge tested in questions is key to completing the GUI tasks. 
For instance, a web browsing interface shows a disabled toggle controlling regional search restriction, with ``Japan'' indicated as the target region. 
The original question in our benchmark asks whether the search results are restricted in japan, which is transformed into its corresponding GUI task ``Set the search results restricted to a specific region: Japan,'' solvable via a single toggle click.
 We define $S1$ as answering the original question correctly, and $S2$ as successfully completing the corresponding GUI tasks. Results are shown in Table \ref{tab:knowledge-correlation}. 
 Prompts for completing GUI tasks are provided in the appendix \cref{guitaskprompt}.
 
\begin{table}[t]

\centering
\caption{Correlation between GUI Knowledge ($S_1$) and Task Completion ($S_2$).}
\resizebox{\linewidth}{!}{
\begin{tabular}{l c c c c}\hline
Model & $P(S_2\checkmark|S_1\checkmark)$ & $P(S_2\times|S_1\checkmark)$ & $P(S_2\times|S_1\times)$ & $P(S_2\checkmark|S_1\times)$ \\ \hline
claude-sonnet-4~\citep{anthropic2025claude4systemcard}  & 20.00\% & 80.00\% & 100.00\% & 0.00\% \\
claude-sonnet-4-5~\citep{anthropic2025claude45} & 8.33\% & 91.67\% & 100.00\% & 0.00\% \\
Doubao-V-Pro~\citep{guo2025seed15vltechnicalreport}  & 0.00\% & 100.00\% & 100.00\% & 0.00\% \\
Gemini-2.5-Pro~\citep{comanici2025gemini25pushingfrontier}  & 0.00\% & 100.00\% & 100.00\% & 0.00\% \\
GPT-5-Chat~\citep{openai2025gpt5systemcard}  & 5.56\% & 94.44\% & 100.00\% & 0.00\% \\\hline
\end{tabular}
}
\vspace{-3mm}
\label{tab:knowledge-correlation}
\end{table}

The results establish GUI knowledge as a necessary but not sufficient condition for task completion: Lacking knowledge guarantees execution failure (e.g., 100\% for Gemini-2.5), validating that knowledge is the strict lower bound for control. Even with correct understanding, execution often fails due to grounding precision. This confirms that our benchmark measures a foundational capability. While having knowledge doesn't guarantee success (due to downstream grounding issues), lacking knowledge guarantees failure.

\section{Conclusion}
We introduces GUI Knowledge Bench, a novel benchmark designed to evaluate the GUI knowledge encoded in vision-language models (VLMs). By analyzing common failure patterns in GUI task execution, the benchmark categorizes GUI knowledge into three dimensions: interface, interaction, and procedure knowledge. The evaluation reveals significant gaps in current VLMs' understanding of system states, interaction effect, and procedure knowledge. These findings highlight the necessity of enriching VLMs with domain-specific GUI knowledge to enhance their performance in real-world GUI tasks and provide insights to guide the development of more capable GUI agents.

\textbf{Impact Statement}
This paper presents work whose goal is to advance the field of machine learning. There are many potential societal consequences of our work, none of which we feel must be specifically highlighted here.

%% file: section/appendix.tex
\subsection{Evaluation Protocol Ablation Studies}
\label{EvaluationProtocol}
\textbf{Option Order Randomization.}

To verify that our multiple-choice evaluation does not rely on positional or linguistic biases, we conduct an ablation study on questions from the interface knowledge subset. We randomly permute all answer options and re-evaluate three representative models.

\begin{table}[t]
\centering
\caption{Effect of option-order randomization on interface knowledge questions. ``Before'' denotes performance on the original option ordering, while ``After'' denotes performance after randomly shuffling all answer options.}
\label{tab:option_shuffle}
\begin{tabular}{lcc}
\hline
Model & Before (\%) & After (\%) \\
\hline
Claude-Sonnet-4 & 79.19 & 80.73 \\
GPT-5-Chat      & 85.16 & 85.93 \\
O3              & 85.16 & 85.74 \\
\hline
\end{tabular}
\end{table}

\cref{tab:option_shuffle} reports model accuracy before and after option shuffling. Performance remains nearly unchanged across all models, indicating that the benchmark is not sensitive to answer ordering. These results suggest that the multiple-choice format does not introduce meaningful bias and that models are not exploiting superficial cues.

\textbf{Free-Form Answer Evaluation.}
To further examine whether models rely on multiple-choice candidates and whether models exploit option cues, we additionally evaluate the questions from interface knowledge subset in a free-form response setting. In this setting, models are required to generate open-ended textual answers without access to predefined options. 
We adopt an LLM-as-a-judge protocol to automatically assess answer correctness. Claude-Sonnet-4, GPT-5-Chat, and O3 are used both as answering models and as independent judges. Results are shown in \cref{tab:free_form}.

\begin{table}[t]
\centering
\caption{Free-form answer evaluation on Interface Knowledge questions using LLM-as-a-judge. Rows indicate the judge model, and columns indicate the evaluation (answering) model.}
\label{tab:free_form}
\resizebox{\linewidth}{!}{
\begin{tabular}{lccc}
\toprule
Judge / Eval Model & Claude-Sonnet-4 & GPT-5-Chat & O3 \\
\midrule
Claude-Sonnet-4 & 47.21 & 52.02 & \textbf{55.11} \\
GPT-5-Chat      & 48.55 & 55.49 & \textbf{57.23} \\
O3              & 47.01 & 53.56 & \textbf{55.88} \\
\bottomrule
\end{tabular}}
\end{table}

Across all models, free-form answering leads to substantially lower accuracy compared to the multiple-choice setting. This performance drop reflects the increased difficulty of recalling and expressing GUI-related knowledge without structural guidance, rather than a change in the underlying knowledge being evaluated.

Importantly, the relative ranking of models remains consistent with the main benchmark results, with O3 achieving higher overall scores. This consistency supports our claim that the benchmark captures stable differences in GUI knowledge and interaction understanding, rather than artifacts introduced by a specific answer format.

\subsection{Question Generation Prompt Template for Interface Knowledge}
\label{messagetemplate}
\textbf{Prompt for widget function.} 

\begin{tcolorbox}[colback=gray!3, colframe=black,
  title=Widget Function Prompt,
  boxrule=0.5pt, arc=2pt, left=2pt, right=2pt, top=4pt, bottom=2pt,
  breakable]

\textcolor{blue!70!black}{\textbf{System Prompt:}} \\

\textbf{[Role]} \\
You will be provided with a single screenshot of a system interface (desktop app, web UI, or mobile app). Generate exactly one challenging GUI reasoning question about that screenshot that requires inspecting the image to answer.

\textbf{[Knowledge Scope of the question]} \\
Ask about the intended function of a specific UI widget (button, toggle, slider, icon, etc.) inferred from the widget's iconography and surrounding context. Avoid universally trivial icons unless combined with contextual clues.

\textbf{[Generation Guidelines]}
\begin{enumerate}
  \item Question length: one concise sentence only. No hints, no steps, no extra context.
  \item Position-only references: Do NOT use any visible text, icon names, or labels from the screenshot. Refer ONLY by position or coordinates (examples: ``top-right corner'', ``third from left in the top toolbar'', ``second row, third column'', ``left sidebar, bottom icon'', or ``\texttt{<x,y>}'' with origin top-left). The question must be unsolvable without the screenshot.
  \item Question types and options:
    \begin{itemize}
      \item If \texttt{multiple\_choice}: produce exactly 4 options. The first option MUST be the correct answer.
      \item If \texttt{yes\_or\_no}: produce exactly 3 options: \{``yes'', ``no'', ``unknown''\} and the correct one must be first.
      \item If the correct answer is genuinely not deducible from the screenshot or you cannot answer the correct answer, then use:
      \begin{itemize}
        \item multiple\_choice: first option = ``none of the other options are correct.''
        \item yes\_or\_no: first option = ``unknown''
      \end{itemize}
    \end{itemize}
  \item Option style: Options must describe actions or effects (not icon shapes). Keep options parallel in length and style ($\approx$ 6--16 words).
  \item Distractors: The 3 incorrect options must be plausible and similar to the correct one.
  \item Contextual reasoning: Prefer questions requiring reasoning across UI elements (e.g., highlighted rows, active tab, enabled/disabled states, adjacent panels).
  \item Based on the provided screenshot, identify which application is currently being used and include this information in your output JSON under the field \texttt{app\_type}.
\end{enumerate}

\textbf{[Output JSON schema --- return exactly this JSON object (no extra text)]}
\begin{json}
{
  "question_type": "multiple_choice" or "yes_or_no",
  "question_text": "<one concise sentence using only positions>",
  "option_text": ["<first option correct>", "<distractor 1>", "<distractor 2>", "<distractor 3>"],
  "app_type": "<application type of the current screenshot>",
  "os_type": "Linux" | "Windows" | "Android" | "MacOS" | "IOS" | "Web"
}
\end{json}

\textbf{[Example Output]}
\begin{json}
{
  "question_type": "yes_or_no",
  "question_text": "While cell B5 in the 'First Name' column shows 'Walter' in the formula bar and the checkmark and 'X' icons are visible beside it, will clicking the 'X' icon clear formatting in the selected cell",
  "option_text": ["yes","no","unknown"],
  "app_type": "Excel",
  "os_type": "Linux"
}
{
  "question_type": "multiple_choice",
  "question_text": "Which of the following statement is correct according to the screenshots?",
  "option_text": [
    "The camera is not currently connected to WiFi",
    "The camera can not be controlled remotely from the phone",
    "Pressing the 'phone' mode icon in the top bar can lead to turning on the phone's airplane mode",
    "Pressing the 'clone' mode icon in the top bar can lead to signing out of the cloud gallery"
  ],
  "app_type": "Excel",
  "os_type": "Linux"
}
\end{json}

\end{tcolorbox}

\textbf{Prompt for layout semantics.} 

\begin{tcolorbox}[colback=gray!3, colframe=black,
  title=Layout Semantics Prompt,
  boxrule=0.5pt, arc=2pt, left=2pt, right=2pt, top=4pt, bottom=2pt,
  breakable]

\textcolor{blue!70!black}{\textbf{System Prompt:}} \\

\textbf{[Role]} \\
You will be provided with a single screenshot of a system interface (desktop app, web UI, or mobile app). Generate exactly one challenging GUI reasoning question about that screenshot that requires inspecting the image to answer.

\textbf{[Knowledge Scope of the question]} \\
The questions should assess whether the model understands positional and grouping relationships between UI elements, inferring their roles from placement and hierarchy. 

\textbf{[Generation Guidelines]}
\begin{enumerate}
  \item Question length: one concise sentence only. No hints, no steps, no extra context.
  \item Position-only references: Do NOT use any visible text, icon names, or labels from the screenshot. Refer ONLY by position or coordinates (examples: ``top-right corner'', ``third from left in the top toolbar'', ``second row, third column'', ``left sidebar, bottom icon'', or ``\texttt{<x,y>}'' with origin top-left). The question must be unsolvable without the screenshot.
  \item Question types and options:
    \begin{itemize}
      \item If \texttt{multiple\_choice}: produce exactly 4 options. The first option MUST be the correct answer.
      \item If \texttt{yes\_or\_no}: produce exactly 3 options: \{``yes'', ``no'', ``unknown''\} and the correct one must be first.
      \item If the correct answer is genuinely not deducible from the screenshot or you cannot answer the correct answer, then use:
      \begin{itemize}
        \item multiple\_choice: first option = ``none of the other options are correct.''
        \item yes\_or\_no: first option = ``unknown''
      \end{itemize}
    \end{itemize}
  \item Option style: Options must describe actions or effects (not icon shapes). Keep options parallel in length and style ($\approx$ 6--16 words).
  \item Distractors: The 3 incorrect options must be plausible and similar to the correct one.
  \item Contextual reasoning: Prefer questions requiring reasoning across UI elements (e.g., highlighted rows, active tab, enabled/disabled states, adjacent panels).
  \item Based on the provided screenshot, identify which application is currently being used and include this information in your output JSON under the field \texttt{app\_type}.
\end{enumerate}

\textbf{[Output JSON schema --- return exactly this JSON object (no extra text)]}
\begin{json}
{
  "question_type": "multiple_choice" or "yes_or_no",
  "question_text": "<one concise sentence using only positions>",
  "option_text": ["<first option correct>", "<distractor 1>", "<distractor 2>", "<distractor 3>"],
  "app_type": "<application type of the current screenshot>",
  "os_type": "Linux" | "Windows" | "Android" | "MacOS" | "IOS" | "Web"
}
\end{json}

\textbf{[Example Output]}
\begin{json}
{
  "question_type": "multiple_choice",
  "question_text": "What is likely to be the departure city?",
  "option_text": ["Beijing", "Shanghai", "Guangzhou", "None of the other options."],
  "app_type": "website",
  "os_type": "Windows"
}

{
  "question_type": "yes_or_no",
  "question_text": "Is the folder in the second row under the 'Documents' folder?",
  "option_text": ["yes", "no", "unknown"],
  "app_type": "Thunderbird",
  "os_type": "Windows"
}

{
  "question_type": "multiple_choice",
  "question_text": "Who sends this email. Please answer the email address.",
  "option_text": ["li@gmail.com", "zhang@gmail.com", "wang@gmail.com", "None of the other options."],
  "app_type": "Email",
  "os_type": "Windows"
}
\end{json}

\end{tcolorbox}


\textbf{Prompt for state information.} 
\begin{tcolorbox}[colback=gray!3, colframe=black,
  title=State Information Prompt,
  boxrule=0.5pt, arc=2pt, left=2pt, right=2pt, top=4pt, bottom=2pt,
  breakable]

\textcolor{blue!70!black}{\textbf{System Prompt:}} \\

\textbf{[Role]} \\
You will be provided with a single screenshot of a system interface (desktop app, web UI, or mobile app). Generate exactly one challenging GUI reasoning question about that screenshot that requires inspecting the image to answer.

\textbf{[Knowledge Scope of the question]} \\
Ask about the current state information of the system, such as whether a control is enabled/disabled, a process is in-progress/completed, a request is pending, or the system is online/offline. Prefer reasoning that requires subtle visual cues or multi-element context.

\textbf{[Generation Guidelines]}
\begin{enumerate}
  \item Question length: one concise sentence only. No hints, no steps, no extra context.
  \item Position-only references: Do NOT use any visible text, icon names, or labels from the screenshot. Refer ONLY by position or coordinates (examples: ``top-right corner'', ``third from left in the top toolbar'', ``second row, third column'', ``left sidebar, bottom icon'', or ``\texttt{<x,y>}'' with origin top-left). The question must be unsolvable without the screenshot.
  \item Question types and options:
    \begin{itemize}
      \item If \texttt{multiple\_choice}: produce exactly 4 options. The first option MUST be the correct answer.
      \item If \texttt{yes\_or\_no}: produce exactly 3 options: \{``yes'', ``no'', ``unknown''\} and the correct one must be first.
      \item If the correct answer is genuinely not deducible from the screenshot or you cannot answer the correct answer, then use:
      \begin{itemize}
        \item multiple\_choice: first option = ``none of the other options are correct.''
        \item yes\_or\_no: first option = ``unknown''
      \end{itemize}
    \end{itemize}
  \item Option style: Options must describe actions or effects (not icon shapes). Keep options parallel in length and style ($\approx$ 6--16 words).
  \item Distractors: The 3 incorrect options must be plausible and similar to the correct one.
  \item Contextual reasoning: Prefer questions requiring reasoning across UI elements (e.g., highlighted rows, active tab, enabled/disabled states, adjacent panels).
  \item Based on the provided screenshot, identify which application is currently being used and include this information in your output JSON under the field \texttt{app\_type}.
\end{enumerate}

\textbf{[Output JSON schema --- return exactly this JSON object (no extra text)]}
\begin{json}
{
  "question_type": "multiple_choice" or "yes_or_no",
  "question_text": "<one concise sentence using only positions>",
  "option_text": ["<first option correct>", "<distractor 1>", "<distractor 2>", "<distractor 3>"],
  "app_type": "<application type of the current screenshot>",
  "os_type": "Linux" | "Windows" | "Android" | "MacOS" | "IOS" | "Web"
}
\end{json}

\textbf{[Example Output]}
\begin{json}
{
  "question_type": "multiple_choice",
  "question_text": "The button in the lower toolbar is active, but the button next to it is greyed out. Which condition is most likely not met yet?",
  "option_text": [
    "All required fields are filled",
    "Network connection is active",
    "File format is supported",
    "None of the other options"
  ],
  "app_type": "Form Editor",
  "os_type": "Web"
}

{
  "question_type": "multiple_choice",
  "question_text": "How can the user enable more controls over the alignment of objects?",
  "option_text": [
    "Select more than one object",
    "Double click the alignment button",
    "None of the other options",
    "User is logged in"
  ],
  "app_type": "Graphics Editor",
  "os_type": "Windows"
}

{
  "question_type": "yes_or_no",
  "question_text": "Will the option in the toolbar become available immediately after selecting a file?",
  "option_text": ["yes","no","unknown"],
  "app_type": "Document Editor",
  "os_type": "MacOS"
}

{
  "question_type": "yes_or_no",
  "question_text": "Is the movie export function currently available?",
  "option_text": ["no","yes","unknown"],
  "app_type": "Video Editor",
  "os_type": "Linux"
}
\end{json}

\end{tcolorbox}



\subsection{Procedure Knowledge Template.}

\begin{tcolorbox}[colback=gray!3, colframe=black,
  title=User Instruction Prompt,
  boxrule=0.5pt, arc=2pt, left=2pt, right=2pt, top=4pt, bottom=2pt,
  breakable]

\textcolor{blue!70!black}{\textbf{User Prompt:}} \\

Analyze the given GUI task and break it down into essential, actionable steps.  
You will receive:  
- a task instruction: \texttt{\{task\_instruction\}}  
- the app where the task occurs: \texttt{\{app\_name\}}  
- the initial screenshot image  

Your goal is to output a Python list of clear, concise steps in logical order to complete the task within the app.  
Each step should represent a key state, action, or milestone.  
Use simple, direct language. Avoid ambiguity or unnecessary complexity.  

\textbf{Output format:}
\begin{itemize}
  \item A valid Python list of strings, e.g.:
\begin{Verbatim}[breaklines,breaksymbolleft={\tiny\ensuremath{\hookrightarrow}\,}]
["First step.", "Second step.", "Third step."]
\end{Verbatim}
  \item Each string must use double quotes ("), and the output must be directly parsable using \texttt{eval()} or \texttt{ast.literal\_eval()}.
  \item Output only the list. No explanation, no extra text.
\end{itemize}

\textbf{Constraints:}
\begin{itemize}
  \item Ensure each step is actionable and unambiguous,
  \item Ensure each step is necessary for task completion,
  \item Ensure each step is easy to follow by a user.
\end{itemize}

\end{tcolorbox}

\subsection{Evaluation Message Prompt Template}
\subsubsection{Interface Knowledge.}
All evaluation questions in this knowledge category use the same prompt template as shown below. 
\begin{tcolorbox}[colback=gray!3, colframe=black,
  title=GUI Agent Inference Prompt,
  boxrule=0.5pt, arc=2pt, left=2pt, right=2pt, top=4pt, bottom=2pt,
  breakable]

\textcolor{blue!70!black}{\textbf{System}} \\
You are a Graphical User Interface (GUI) agent. You will be given a screenshot, a question, and corresponding options. You need to choose one option as your answer.

\vspace{4pt}
\textcolor{blue!70!black}{\textbf{User}} \\
\texttt{\{question\_images\}} \\
\texttt{\{question\_texts\}} \\
\texttt{\{question\_options\}}

\vspace{6pt}
\textcolor{blue!70!black}{\textbf{Response Rules}} \\

\textbf{If question\_type == 'yes\_or\_no'}: \\
Think step by step. You must respond strictly in JSON format following this schema:
\begin{json}
{
  "thought": "<your reasoning>",
  "answer": "<yes/no/unknown>"
}
\end{json}

\textbf{If question\_type == 'multiple\_choice'}: \\
Think step by step. You must respond strictly in JSON format following this schema:
\begin{json}
{
  "thought": "<your reasoning>",
  "answer": "<A/B/C/D>"
}
\end{json}

\end{tcolorbox}

\textbf{Interaction Knowledge.}

\begin{tcolorbox}[colback=gray!3, colframe=black,
  title=GUI Agent Task-Solving Prompt,
  boxrule=0.5pt, arc=2pt, left=2pt, right=2pt, top=4pt, bottom=2pt,
  breakable]

\textcolor{blue!70!black}{\textbf{System}} \\
You are a Graphical User Interface (GUI) agent. You will be given a task instruction, a screenshot, several GUI operations, and four options. Your goal is to select the best option that could solve the task. \\
\texttt{\{question\_images\}}

\vspace{4pt}
\textcolor{blue!70!black}{\textbf{User}} \\
\texttt{\{question\_text\}} \\
Which of the above options are correct according to the screenshots? Think step by step. You must respond strictly in JSON format following this schema.

\vspace{6pt}
\textcolor{blue!70!black}{\textbf{Response Schema}} \\
\begin{json}
{
  "thought": "<your reasoning>",
  "answer": "<A/B/C/D>"
}
\end{json}

\end{tcolorbox}

\subsubsection{Interaction Knowledge}
\textbf{Interaction Effect}

\begin{tcolorbox}[colback=gray!3, colframe=black,
  title=GUI Agent Next-State Selection Prompt,
  boxrule=0.5pt, arc=2pt, left=2pt, right=2pt, top=4pt, bottom=2pt,
  breakable]

\textcolor{blue!70!black}{\textbf{System}} \\
You are a Graphical User Interface (GUI) agent. You will be given a screenshot, action descriptions, and multiple options, each containing an image. After performing one action on the screenshot, your goal is to select the option that correctly corresponds to the resulting screenshot after performing the action. Below is a short description of the action space:

\begin{Verbatim}[breaklines,breaksymbolleft={\tiny\ensuremath{\hookrightarrow}\,},fontsize=\small]
if platform == Desktop:
        Action Space
        - click(point='x1 y1'): left click a position on the screen. 
        - left_double(point='x1 y1'): left double click a position on the screen. 
        - right_single(point='x1 y1'): right single click a position on the screen. 
        - drag(start_point='x1 y1', end_point='x2 y2'): drag the mouse from one position to another. 
        - hotkey(key='ctrl c'): keyboard shortcut, split keys with spaces
        - type(content='xxx'): type an answer, use escape characters (', ", \n) when needed. Add \n at the end if it is the final submission.
        - scroll(point='x1 y1', direction='down or up or right or left'): scroll to see more content
        
if platform == Mobile:
        Action Space
        - click(point='x1 y1')
        - long_press(point='x1 y1')
        - type(content='') #If you want to submit your input, use "\\n" at the end of `content`.
        - scroll(point='x1 y1', direction='down or up or right or left'): scroll to see more content
\end{Verbatim}

The size of the image is \texttt{\{w\}x\{h\}}.\ \texttt{\textbackslash n}

\vspace{4pt}
\textcolor{blue!70!black}{\textbf{User}} \\
\texttt{\{question\_image\}} \\
Above is the current screenshot. \\
After I perform the described action \texttt{'action\_type(action\_parameter)'} (as drawn in the initial screenshot), which of the following options correctly corresponds to the resulting screenshot?

A. \texttt{\{option\_image\_A\}} \\
B. \texttt{\{option\_image\_B\}} \\
C. \texttt{\{option\_image\_C\}} \\
D. \texttt{\{option\_image\_D\}}

\vspace{6pt}
\textcolor{blue!70!black}{\textbf{Response Schema}} \\
Think step by step. You must respond strictly in JSON format following this schema:
\begin{json}
{
  "thought": "<your reasoning>",
  "answer": "<A/B/C/D>"
}
\end{json}

\end{tcolorbox}

\textbf{Interaction Parameter}

\begin{tcolorbox}[colback=gray!3, colframe=black,
  title=GUI Agent Action-Parameter Selection Prompt,
  boxrule=0.5pt, arc=2pt, left=2pt, right=2pt, top=4pt, bottom=2pt,
  breakable]

\textcolor{blue!70!black}{\textbf{System}} \\
You are a Graphical User Interface (GUI) agent. You will be given two consecutive screenshots of the GUI, action descriptions, and multiple options. Your goal is to select which action was performed to transition from the first screenshot to the second. If the description specifies an action type, select the correct parameter value for the given action.

\begin{Verbatim}[breaklines,breaksymbolleft={\tiny\ensuremath{\hookrightarrow}\,},fontsize=\small]
if platform == Desktop:
        Action Space
        - click(point='x1 y1'): left click a position on the screen. 
        - left_double(point='x1 y1'): left double click a position on the screen. 
        - right_single(point='x1 y1'): right single click a position on the screen. 
        - drag(start_point='x1 y1', end_point='x2 y2'): drag the mouse from one position to another. 
        - hotkey(key='ctrl c'): keyboard shortcut, split keys with spaces
        - type(content='xxx'): type an answer, use escape characters (', ", \n) when needed. Add \n at the end if it is the final submission.
        - scroll(point='x1 y1', direction='down or up or right or left'): scroll to see more content
        
if platform == Mobile:
        Action Space
        - click(point='x1 y1')
        - long_press(point='x1 y1')
        - type(content='') #If you want to submit your input, use "\\n" at the end of `content`.
        - scroll(point='x1 y1', direction='down or up or right or left'): scroll to see more content
\end{Verbatim}

The size of the image is \texttt{\{w\}x\{h\}}.\ \texttt{\textbackslash n}

\texttt{\{question\_images\}}

\vspace{6pt}
\textcolor{blue!70!black}{\textbf{User}} \\
Above are two consecutive screenshots. Your task is to select the option containing the right parameter value of the given action \texttt{' \{action\_type\} '} to transition from the first to the second screenshot. \\
As is drawn in the first screenshot. Which of the above options are correct according to the screenshots?

A. \texttt{\{option\_text\}} \\
B. \texttt{\{option\_text\}} \\
C. \texttt{\{option\_text\}} \\
D. \texttt{\{option\_text\}}

\vspace{6pt}
\textcolor{blue!70!black}{\textbf{Response Schema}} \\
Think step by step. You must respond strictly in JSON format following this schema:
\begin{json}
{
  "thought": "<your reasoning>",
  "answer": "<A/B/C/D>"
}
\end{json}

\end{tcolorbox}

\textbf{Interaction Type}

\begin{tcolorbox}[colback=gray!3, colframe=black,
  title=GUI Agent Action Identification Prompt,
  boxrule=0.5pt, arc=2pt, left=2pt, right=2pt, top=4pt, bottom=2pt,
  breakable]

\textcolor{blue!70!black}{\textbf{System}} \\
You are a Graphical User Interface (GUI) agent. You will be given two consecutive screenshots of the GUI, action descriptions, and multiple options. Your goal is to select which action was performed to transition from the first screenshot to the second. If the description specifies an action type, select the correct parameter value for the given action.

\begin{Verbatim}[breaklines,breaksymbolleft={\tiny\ensuremath{\hookrightarrow}\,},fontsize=\small]
if platform == Desktop:
        Action Space
        - click(point='x1 y1'): left click a position on the screen. 
        - left_double(point='x1 y1'): left double click a position on the screen. 
        - right_single(point='x1 y1'): right single click a position on the screen. 
        - drag(start_point='x1 y1', end_point='x2 y2'): drag the mouse from one position to another. 
        - hotkey(key='ctrl c'): keyboard shortcut, split keys with spaces
        - type(content='xxx'): type an answer, use escape characters (', ", \n) when needed. Add \n at the end if it is the final submission.
        - scroll(point='x1 y1', direction='down or up or right or left'): scroll to see more content
        
if platform == Mobile:
        Action Space
        - click(point='x1 y1')
        - long_press(point='x1 y1')
        - type(content='') #If you want to submit your input, use "\\n" at the end of `content`.
        - scroll(point='x1 y1', direction='down or up or right or left'): scroll to see more content
\end{Verbatim}

The size of the image is \texttt{\{w\}x\{h\}}.\ \texttt{\textbackslash n}

\texttt{\{question\_images\}}

\vspace{6pt}
\textcolor{blue!70!black}{\textbf{User}} \\
Above are two consecutive screenshots. Your task is to select which action is performed in order to transition from the first screenshot to the second.

\begin{Verbatim}[breaklines,breaksymbolleft={\tiny\ensuremath{\hookrightarrow}\,},fontsize=\small]
if platform == Desktop:
    {seven action types}
    Which of the above options are correct according to the screenshots?
    Think step by step. You must respond strictly in JSON format following this schema:
    {"thought": "<your reasoning>", "answer": "<A/B/C/D/E/F/G>" }

if platform == Mobile:
    {four action types}
    Which of the above options are correct according to the screenshots?
    Think step by step. You must respond strictly in JSON format following this schema:
    {"thought": "<your reasoning>", "answer": "<A/B/C/D>" }
\end{Verbatim}

\vspace{6pt}
\textcolor{blue!70!black}{\textbf{Response Schema (Desktop)}} \\
\begin{json}
{
  "thought": "<your reasoning>",
  "answer": "<A/B/C/D/E/F/G>"
}
\end{json}

\textcolor{blue!70!black}{\textbf{Response Schema (Mobile)}} \\
\begin{json}
{
  "thought": "<your reasoning>",
  "answer": "<A/B/C/D>"
}
\end{json}

\end{tcolorbox}

\subsubsection{Procedure Knowledge}
\textbf{Task Objective}

\begin{tcolorbox}[colback=gray!3, colframe=black,
  title=Task Completion Verification Prompt,
  boxrule=0.5pt, arc=2pt, left=2pt, right=2pt, top=4pt, bottom=2pt,
  breakable]

\textcolor{blue!70!black}{\textbf{System}} \\
You are a Graphical User Interface (GUI) agent. You will be given a sequence of screenshots, a task instruction, and three possible answer options: \texttt{yes}, \texttt{no}, \texttt{unknown}. Your goal is to select the best option that indicates whether the task is completed.

\begin{itemize}
  \item \textbf{yes} --- The task is clearly completed.
  \item \textbf{no} --- The task is not completed.
  \item \textbf{unknown} --- The screenshots do not provide enough evidence to determine completion.
\end{itemize}

\vspace{4pt}
\textcolor{blue!70!black}{\textbf{User}} \\
According to the screenshots below, has the task "\texttt{\{task\}}" been completed? \\
\texttt{\{question\_images\}}

\vspace{6pt}
\textcolor{blue!70!black}{\textbf{Response Schema}} \\
Think step by step. You must respond strictly in JSON format following this schema:
\begin{json}
{
  "thought": "<your reasoning>",
  "answer": "<yes/no/unknown>"
}
\end{json}

\end{tcolorbox}

\textbf{Workflow Sequence}

\begin{tcolorbox}[colback=gray!3, colframe=black,
  title=GUI Agent Conditional QA Prompt,
  boxrule=0.5pt, arc=2pt, left=2pt, right=2pt, top=4pt, bottom=2pt,
  breakable]

\textcolor{blue!70!black}{\textbf{System}} \\

\textbf{If \texttt{question\_type == 'yes\_or\_no'}:} \\
You are a Graphical User Interface (GUI) agent. You will be given a screenshot, a question, and corresponding options. You need to choose one option as your answer.

\medskip
\textbf{If \texttt{question\_type == 'multiple\_choice'}:} \\
You are a Graphical User Interface (GUI) agent. You will be given a task instruction, a screenshot, several GUI operations, and four options. Your goal is to select the best option that could solve the task.

\medskip
\texttt{\{question\_images\}}

\vspace{6pt}
\textcolor{blue!70!black}{\textbf{User}} \\
\texttt{\{question\_text\}} \\
\texttt{\{option\_texts\}} \\
Which of the above options are correct according to the screenshot?

\vspace{6pt}
\textcolor{blue!70!black}{\textbf{Response Rules}} \\

\textbf{If \texttt{question\_type == 'yes\_or\_no'}:} \\
Think step by step. You must respond strictly in JSON format following this schema:
\begin{json}
{
  "thought": "<your reasoning>",
  "answer": "<yes/no/unknown>"
}
\end{json}

\textbf{If \texttt{question\_type == 'multiple\_choice'}:} \\
Think step by step. You must respond strictly in JSON format following this schema:
\begin{json}
{
  "thought": "<your reasoning>",
  "answer": "<A/B/C/D>"
}
\end{json}

\end{tcolorbox}


\subsection{Prompts for completing GUI tasks}
\label{guitaskprompt}
\begin{tcolorbox}[colback=gray!3, colframe=black,
  title=GUI Agent Conditional QA Prompt,
  boxrule=0.5pt, arc=2pt, left=2pt, right=2pt, top=4pt, bottom=2pt,
  breakable]

\textcolor{blue!70!black}{\textbf{System}} \\

You are a helpful assistant.

\textcolor{blue!70!black}{\textbf{User}} \\

You are a GUI agent. You are given a task and your action history, with screenshots. You need to perform the next action to complete the task.

\medskip
\textbf{Output Format} \\
```
    Thought: ...
    Action: ...
 ```

\medskip
\textbf{ Action Space} \\
click((x1,y1))\\
left\_double((x1,y1))\\
right\_single((x1,y1))\\
drag((x1,y1), (x2,y2))\\
hotkey(key='')\\
type(content='')  If you want to submit your input, use "\\n" at the end of `content`.\\
scroll((x1,y1), direction='down or up or right or left')\\
done()  If you think the instruction is finished, parameters none\\

\medskip
\textbf{ User Instruction} \\
    Task Instruction

\medskip
\textbf{Image Info} \\
Image size (pixels): width={image\_size[0]}, height={image\_size[1]}. Output absolute pixel coordinates.

\end{tcolorbox}

\subsection{Dataset Statistics Overview}
\label{datasetstatistics}
\begin{figure*}
    \centering
    \includegraphics[width=\linewidth]{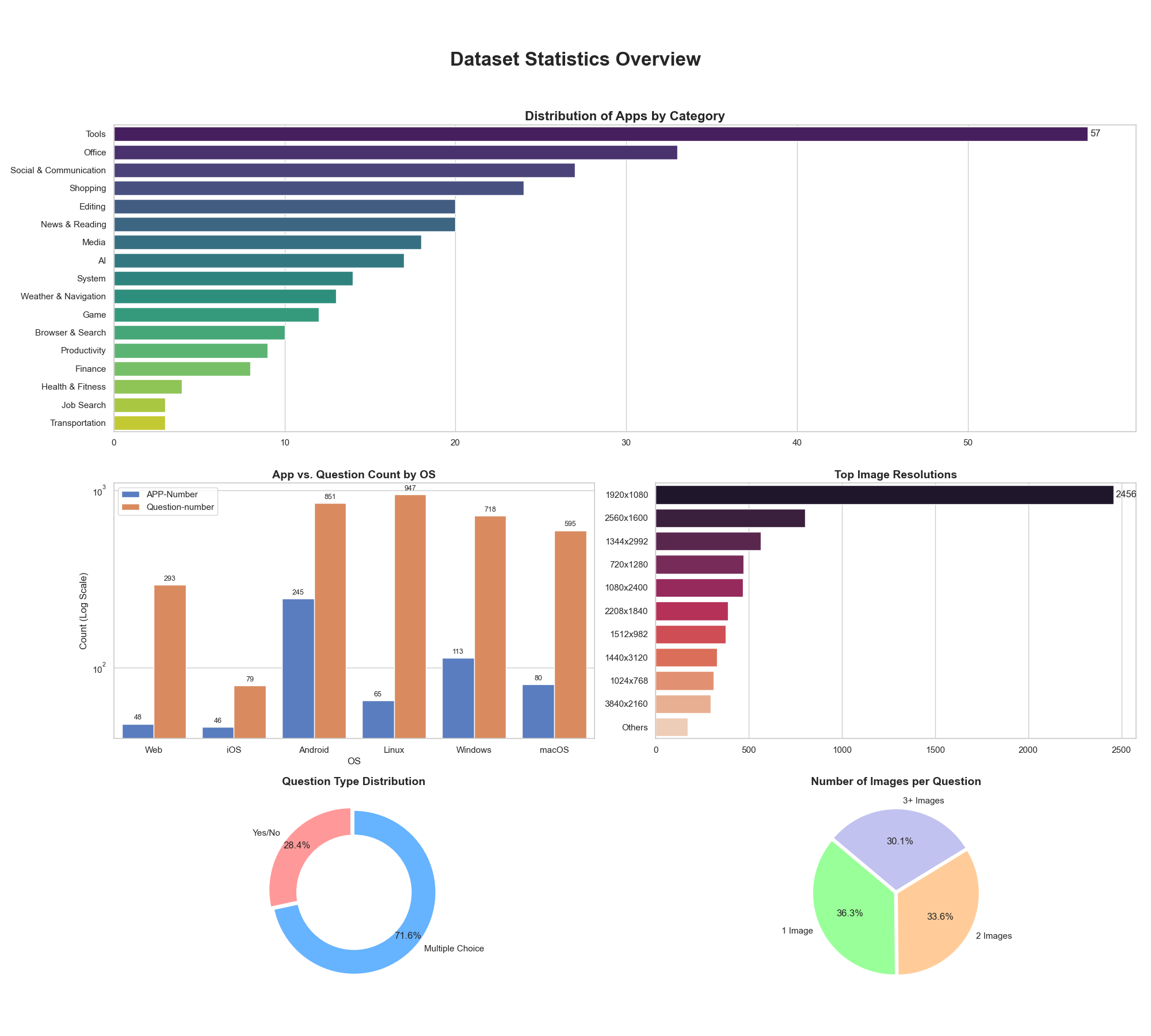}
    \caption{Dataset Statistics Overview}
    \label{fig:Dataset Statistics Overview}
\end{figure*}

We show more detailed statistics of our benchmark in Figure \ref{fig:Dataset Statistics Overview}. 

\subsection{Full Application List}
Here we include the full list of applications involved in our benchmark. 
\begin{tcolorbox}[colback=gray!3, colframe=black,
  title=List of Applications,
  boxrule=0.5pt, arc=2pt, left=2pt, right=2pt, top=4pt, bottom=2pt,
  breakable]

\textbf{Office (30)}: Apple Notes, Apple Reminders, Calendar, Docs, Document Viewer, Evince, Gedit, Google Calendar, Google Docs, Google Keep, Keynote, Lark, Libreoffice, Notability, Note-taking App, Notepad, Notes, Notion, Numbers, Office, Overleaf, Pages, Powerpoint, Spreadsheet, Text Editor, VS Code, WPS Office, Microsoft Word, Xcode, Freeform.

\medskip
\textbf{Media (18)}: Amazon Music, Amazon Prime Video, Iheartradio, Likee, Music, Music Player, Pandora, Pocket FM, Podcast Player, Quicktime, Roku, Sofascore, Spotify, TikTok, Tubi, VLC media player, YouTube, YouTube Music.

\medskip
\textbf{Game (12)}: Arena\_of\_valor, CS2, Chess, Defense\_of\_the\_ancients\_2, Dream, Genshin\_impact, Minecraft, Nintendo, Pubg, Red\_dead\_redemption\_2, Steam, The Legend Of Zelda Breath Of The Wild.

\medskip
\textbf{Editing (20)}: 3dviewer, Adobe Acrobat, Adobe After Effects, Adobe Express, Adobe Photoshop, Adobe Photoshop Express, Adobe Premiere Pro, CapCut, Davinci Resolve, Draw.io, Gimp, Paint, PDF Editor, Photo Editing Tool, Photo Editor, Picsart, Procreate, Runway, Snapseed, Video Editing Software.

\medskip
\textbf{Social \& Communication (28)}: Discord, Facebook, Flickr, Gmail, Google Meet, Google Messages, Imessage, Instagram, LinkedIn, Mail, Messenger, Outlook, Phone, Pinterest, Quora, Reddit, Signal, Slack, Teams Live, Telegram, Threads, Thunderbird, Tumblr, WeChat, Weibo, WhatsApp, X (Twitter), Zoom.

\medskip
\textbf{Shopping (25)}: 12306, Alibaba, Aliexpress, Amazon Shopping, Apartments.com, Applestore, Autoscout24, Autouncle, Booking.com, Car Marketplace, Cars.co.za, Ebay, Edmunds, Expedia, Magento, Offerup, Onestopmarket, Product Listing App, Realtor.com, Redfin, Shop, Taobao, Tripadvisor, Walmart, Wish.

\medskip
\textbf{AI \& Tools (17)}: AI Art Generator, Align-anything-dev-omni, Amazon Alexa, Chatbot AI, Chatgpt, Chaton AI, DeepL Translate, Google Translate, Grammarly, Microsoft Copilot, Microsoft Translator, Remix AI Image Creator, Stable Diffusion, Translate, WOMBO Dream, Yandex Translate, Zhiyun Translate.

\medskip
\textbf{Browser \& Search (10)}: Bing, DuckDuckGo, Firefox, Google App, Google Chrome, Google Search, Opera, Safari, Web Browser, Web.

\medskip
\textbf{Tools (60)}: Accerciser, Activities, Activity Monitor, App Lock, App Locker, Applock Pro, Automator, Baidu Netdisk, Bluetoothnotificationareaiconwindowclass, Calculator, Camera, Clean, ClevCalc - Calculator, Color Management Utility, Colorsync\_utility, Contacts, Control Center, Cursor, Desktop, Dictionary, Digital Color Meter, Disk Utility, Drops, Electron, Email Client, File, File Explorer, File Manager, Files, Filezilla, Finder, Font Book, GPS, Image Viewer, Iphonelockscreen, Kid3, Launcher, Mi Mover, Microsoft Store, Preview, Recorder, Rosetta Stone, Scientific Calculator Plus 991, Script\_editor, Search, Shortcuts, Spotlight, Stickies, System Information, System Search, System Settings, Task Manager, Terminal, Totem, ToDesk, Trash, Vim, Voicememos, Vottak, Wallpaper Picker.

\medskip
\textbf{Productivity (9)}: Any.do, Drive, Dropbox Paper, Google Drive, Onedrive, Paperflux, Things, TickTick, Todoist.

\medskip
\textbf{News \& Reading (22)}: AP News, BBC News, BBC Sport, Bloomberg, Crimereads, Espn, Forbes, Goodreads, Google News, Google Play Books, Google Scholar, Kindle, Kobo Books, Metacritic, Microsoft News, Newsbreak, Wikidata, Wikipedia, Yahoo Sports, Apple News, Travel Guide App, Travel Review App.

\medskip
\textbf{Weather \& Navigation (12)}: Accuweather, Apple Maps, Citymapper, Google Maps, Mapillary, Miuiweather, Msnweather, Navigation App, Openstreetmap, Waze, Weather, Windy.

\medskip
\textbf{Finance (8)}: Alipay, Budgeting App, Investing.com, Paymore, Stocks, Wallet For Your Business, Wallet: Budget Money Manager, Yahoo Finance.

\medskip
\textbf{Health \& Fitness (4)}: Fitbit, Fiton, Mideaair, Mifitness.

\medskip
\textbf{Job Search (3)}: Indeed, Job Search By Ziprecruiter, Ziprecruiter.

\medskip
\textbf{Transportation (3)}: Didi, Ryanair, Uber.

\medskip
\textbf{System \& Tools (15)}: Android, Android Home Screen, Android Launcher, Android Settings, Android Share Sheet, App Store, Apple, Applibrary, Gnome, Mobile Home Launcher, Mobile Launcher, Mobile Web Browser, OS, Ubuntu, Ubuntu Desktop.

\end{tcolorbox}

\subsection{Action Type Prediction Confusion Matrix}
\label{actionconfusion}
\begin{figure}
    \centering
    \includegraphics[width=\linewidth]{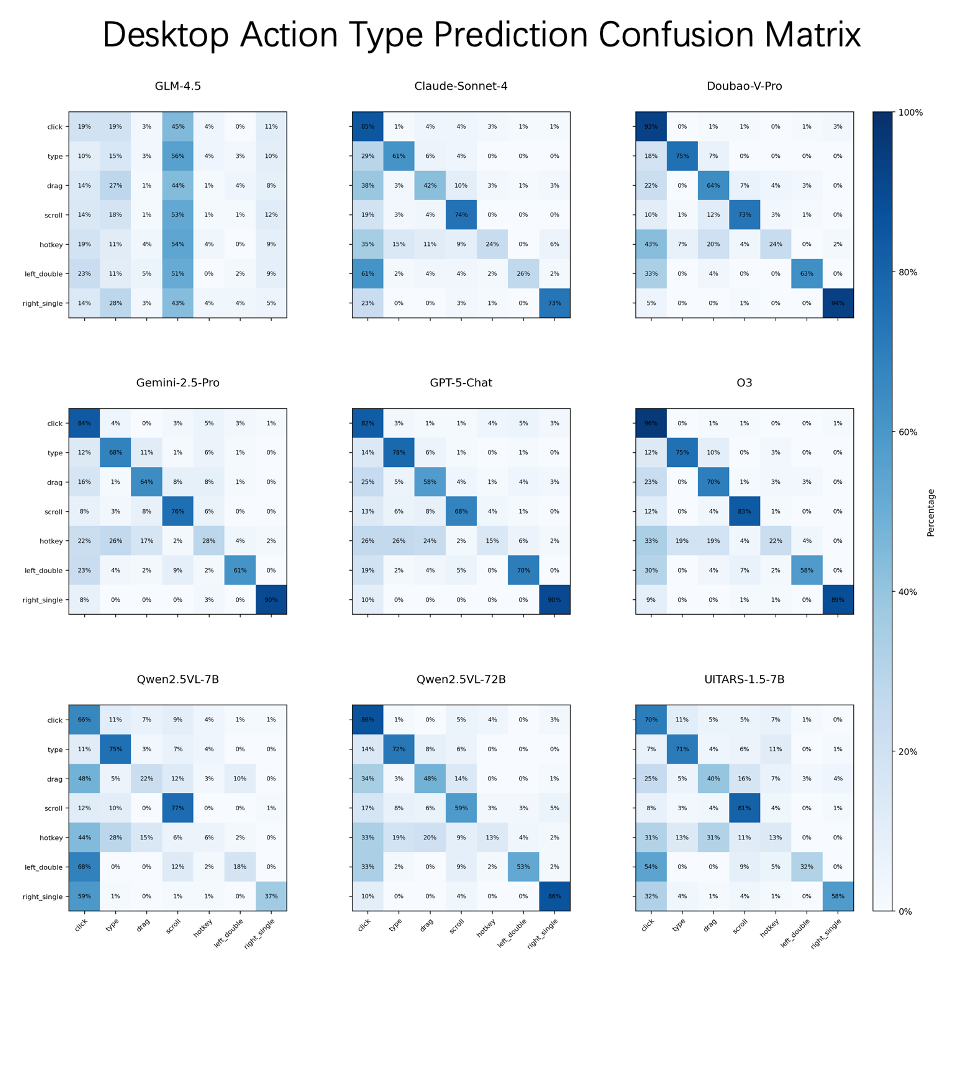}
    \caption{Confusion matrix of action type prediction in desktop.}
    \label{fig:Confusion}
\end{figure}

\begin{figure}
    \centering
    \includegraphics[width=\linewidth]{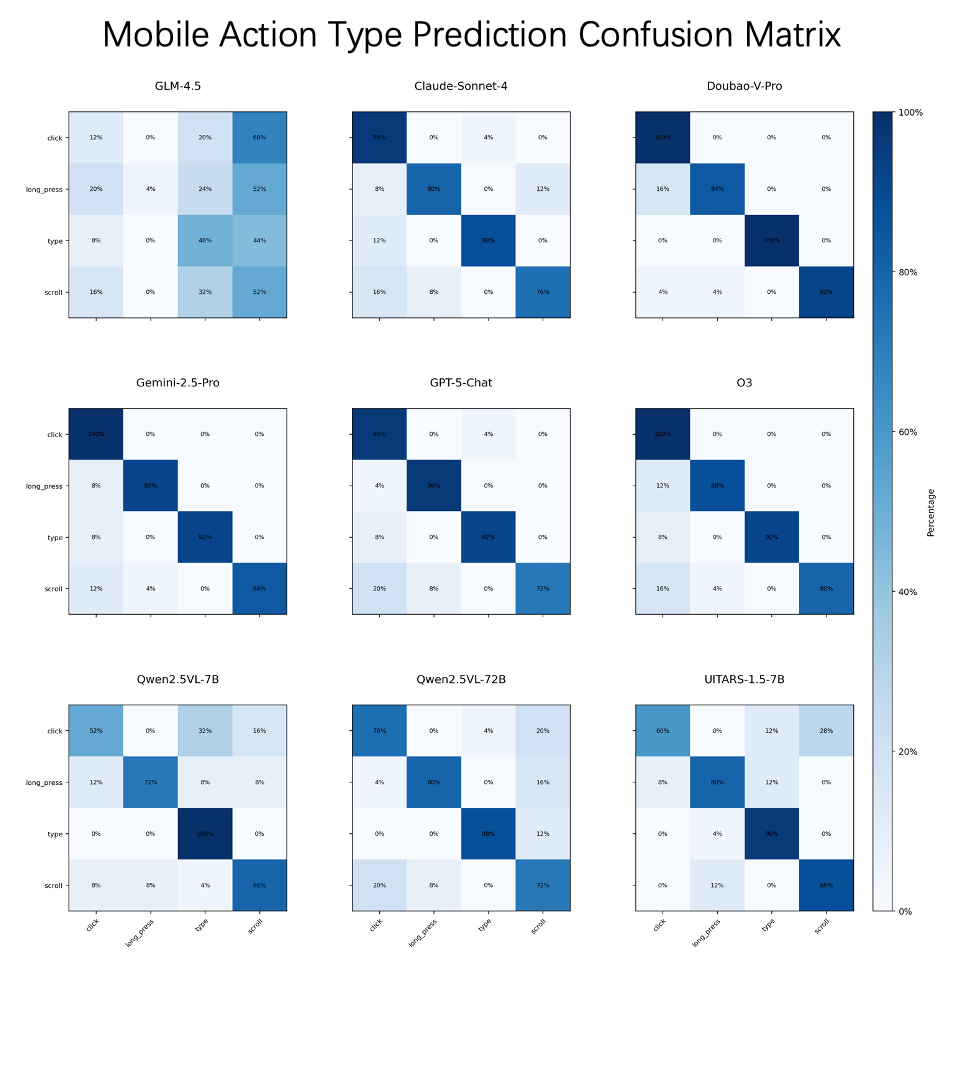}
    \caption{Confusion matrix of action type prediction in mobile.}
    \label{fig:Confusion2}
\end{figure}

Figure~\ref{fig:Confusion} and Figure~\ref{fig:Confusion2} show the confusion matrix of tested models on desktop and mobile. All of these models have a tendency for predicting click instead of the right actions. 

